\documentclass[11pt,draftclsnofoot,onecolumn,journal,letterpaper]{IEEEtran}

\usepackage{graphicx}
\usepackage{subfigure}
\usepackage{comment}
\usepackage[cmex10]{amsmath}
\usepackage{amsfonts}
\usepackage{amssymb,amsthm}
\usepackage{algorithm}
\usepackage{algorithmic}
\usepackage{cite}
\usepackage{bm}
\usepackage{mathrsfs}
\usepackage{color}
\usepackage{multirow}
\usepackage{epstopdf}
\usepackage{setspace}

\theoremstyle{definition}
\newtheorem{remark}{Remark}

\newcommand{\tabincell}[2]{\begin{tabular}{@{}#1@{}}#2\end{tabular}}
\newcommand{\ssf}[1]{\textrm{$\sf{#1}$}{}}

\begin{document}

\title{An Iterative BP-CNN Architecture for \\Channel Decoding}

\author{Fei~Liang, 
	Cong~Shen, 
        and~Feng~Wu
\thanks{This work has been supported by the National Natural Science Foundation of China under Grant 61631017.}
\thanks{The authors are with the Laboratory of Future Networks, School of Information Science and Technology, University of Science and Technology of China. }
}

\maketitle

\begin{abstract}

Inspired by recent advances in deep learning, we propose a novel iterative BP-CNN architecture for channel decoding under correlated noise. This architecture concatenates a trained convolutional neural network (CNN) with a standard belief-propagation (BP) decoder. The standard BP decoder is used to estimate the coded bits, followed by a CNN to remove the estimation errors of the BP decoder and obtain a more accurate estimation of the channel noise. Iterating between BP and CNN will gradually improve the decoding SNR and hence result in better decoding performance. To train a well-behaved CNN model, we define a new loss function which involves not only the accuracy of the noise estimation but also the normality test for the estimation errors, i.e., to measure how likely the estimation errors follow a Gaussian distribution. The introduction of the normality test to the CNN training shapes the residual noise distribution and further reduces the BER of the iterative decoding, compared to using the standard quadratic loss function. We carry out extensive experiments to analyze and verify the proposed framework. The iterative BP-CNN decoder has better BER performance with lower complexity, is suitable for parallel implementation, does not rely on any specific channel model or encoding method, and is robust against training mismatches. All of these features make it a good candidate for decoding modern channel codes.

\end{abstract}

\IEEEpeerreviewmaketitle

\section{Introduction}
Channel encoding and decoding are important components in modern communication systems. Inspired by Shannon's original work \cite{shannon2001mathematical}, tremendous progresses have been made both in coding theory and its applications. For example, low-density parity-check (LDPC) codes \cite{gallager1962low} are able to yield a performance close to the Shannon capacity for AWGN channels with a properly optimized encoding structure and the well developed belief-propagation (BP) decoding algorithm \cite{Richardson2001}. As another example, short to medium-block length linear codes  such as Bose-Chaudhuri-Hocquenghem (BCH) code \cite{LC:04} and Polar code \cite{Arikan09} are of great interest to the recent development in 5G \cite{Liva2016}, for delay-sensitive and mission-critical applications. 


However, in practical communication systems, channels sometimes exhibit correlations in the noise samples due to filtering, oversampling \cite{sharma2013snr}, channel fading and multi-user interference \cite{Honig2009}. A well-designed channel code may not have satisfactory performance if the receiver is not designed to handle noise correlations. The difficulty in addressing this issue mainly comes from the high-complexity model introduced by the colored noise. In theory, the decoder can first estimate the noise distribution, and then optimize the BP decoder using the estimated joint distribution. This approach, nevertheless, is model-based and may suffer from not obtaining a well-behaved  joint distribution for noise samples. Furthermore, optimizing the BP decoder with a joint noise distribution may be highly complex when the correlation is strong. Hence, a low-complexity, highly-adaptive and robust decoder structure that can fully exploit the characteristics of noise correlations and have good decoding performance is desired.


Recent advances in deep learning  provide a new direction to tackle this problem. Instead of finding an algorithm based on a pre-defined model,Πdeep learning technologies allow the system to learn an efficient algorithm directly from training data. Deep learning has been applied in computer vision \cite{he2016deep}, natural language processing \cite{sutskever2014sequence}, autonomous vehicles \cite{chen2015deepdriving} and many other areas, and the results have been quite remarkable.  Inspired by these advances, researchers have recently tried to solve communication problems (including channel decoding) using deep learning technologies \cite{nachmani2016learning,o2016learning,o2016end,nachmani2017rnn,lugosch2017neural,gruber2017deep,cammerer2017scaling,o2017introduction,farsad2017detection,samuel2017deep,Dorner2017}, and a summary of these works is provided in Section~\ref{sec:related_work}. However, none of these works address the problem of efficient  decoding of linear codes under correlated channel noise.


In this paper, we design a novel receiver architecture to tackle the decoding problem when correlation exists in channel noise. This architecture concatenates a trained convolutional neural network (CNN) with a standard BP decoder, and the received symbols are iteratively processed between BP and CNN -- hence the name \textit{iterative BP-CNN}. At the receiver side, the received symbols are first processed by the BP decoder to obtain an initial decoding. Then, subtracting the estimated transmit symbols from the received symbols, we obtain an estimation of channel noise. Because of the decoding error, the channel noise estimation is not exact. We then feed the channel noise estimation into a CNN, which further removes the estimation errors of the BP decoder and obtains a more accurate noise estimation  by exploiting the noise correlation via training. Iterating between BP and CNN will gradually improve the decoding SNR and hence result in better decoding performance. 

The proposed  iterative BP-CNN decoder has many desirable properties. As we will see, it has better decoding performance than the standard BP, with lower complexity. This is mainly due to the efficient CNN structure, which consists mostly of linear operations and a few non-linear ones. This method is data driven and does not rely on any pre-defined models, which makes it adaptive to different channel conditions and different linear codes. It is also suitable for parallel computing, which is  important for VLSI implementation.  Furthermore, our experiments show that the iterative BP-CNN decoder is robust to SNR mismatches if the training data is generated carefully.

Intuitively, the reason that CNN can help channel decoding is similar to the success of CNN in low-level tasks in image processing such as image denoising \cite{jain2009natural} or image super-resolution \cite{dong2016image}. This becomes more clear when we view the correlation in channel noise as a ``feature'', which can be extracted by CNN. However, our problem setting is very different to these other applications where extracting features is the ultimate goal. In the iterative BP-CNN architecture, the goal of CNN is not only to accurately estimate the channel noise and depress the residual error, but also to generate an output that is beneficial to the BP decoder. This unique requirement has motivated us to develop a novel loss function for CNN training, which combines the effect of residual noise power with a Jarque-Bera normality test. We will see that this new loss function plays an important role in the \emph{enhanced BP-CNN} decoder.

In summary, our contributions in this work are as follows:
\begin{enumerate}
	\item We propose a novel decoding structure for linear codes, called \textit{iterative BP-CNN}, that concatenates a trained CNN with a standard BP decoder and iterates between them. This structure is shown to have the capability to extract noise correlation features and improve the decoding performance. 
	\item We design the CNN architecture for \textit{iterative BP-CNN} with two different loss functions. In the \emph{baseline BP-CNN}, the well known quadratic loss function is used in CNN training. In the  \emph{enhanced BP-CNN}, we develop a new loss function that depresses the residual noise power and simultaneously performs the Jarque-Bera normality test. 
	\item We carry out extensive evaluations to verify and analyze the proposed framework.
\end{enumerate}

The rest of this paper is organized as follows. In Section \ref{sec:related_work} we give a brief review of some related works. The system design, including the network training, is explained in detail in Section \ref{sec:system_design}. Extensive experiments is in Section \ref{sec:experiment}. We finally conclude this paper and discuss future works in Section \ref{sec:conclusion}.

\section{Related works}  
\label{sec:related_work}

\subsection{Convolutional Neural Networks} 
Recent progresses of deep learning technologies, big data, and powerful computational devices such as GPUs have started a new era in artificial intelligence. Taking computer vision as an example, deep convolutional neural networks have been verified to yield much better performance than conventional methods in various applications, from high-level tasks such as image recognition \cite{krizhevsky2012imagenet,he2016deep} and object detection \cite{ren2015faster}, to low-level tasks such as image denoising \cite{jain2009natural} and image super-resolution \cite{dong2016image}. In \cite{krizhevsky2012imagenet}, the authors propose to use a deep CNN for image classification, consisting of convolutional layers, pooling layers, and fully-connected layers. A variant of this model won the ILSVRC-2012 competition, and CNNs have received skyrocketed interest in both academia and industry since then.  \cite{he2016deep}, the authors have designed a residual learning framework (ResNet) which makes it easier to train deeper neural networks. It is shown that ResNet can even outperform humans in some high-level tasks. With regard to low-level tasks such as image super-resolution, the authors of \cite{dong2016image} show that a fully convolutional neural network can achieve similar restoration quality with the-state-of-art methods but with low complexity for practical applications. More recently, CNNs have played a crucial role in AlphaGo \cite{silver2016mastering} which beats the best human player in Go games.

The task of channel decoding under colored noise is similar to some low-level tasks such as image denoising or super-resolution, which has motivated us to combine a CNN with BP in this work. We will elaborate on this similarity in Section~\ref{sec:system_design}.

\subsection{Applications of Deep Learning in Communications} 
Most algorithms in communications are developed based on given models. For some tasks, this is only feasible when the model is simple enough, and model mismatch often happens in practice. Recent advances in deep learning technologies have attracted attention of researchers in communications and there have been some recent papers on solving communication problems using deep learning technologies. For the topic of designing channel decoding algorithms, which is the focus of this paper, Nachmani \textit{et al.} have demonstrated that by assigning proper weights to the passing messages in the Tanner graph, comparable decoding performance can be achieved with less iterations than traditional BP decoders \cite{nachmani2016learning}. These weights are obtained via training in deep learning, which partially compensate for the effect of small cycles in the Tanner graph. The authors further introduce the concept of recurrent neural networks (RNN) into BP decoding in \cite{nachmani2017rnn}. Combined with the modified random redundant (mRRD) iterative algorithm \cite{dimnik2009improved}, additional performance improvement can be obtained. Considering that BP decoding contains many multiplications, Lugosch \textit{et al.} have proposed a light-weight neural offset min-sum decoding algorithm in \cite{lugosch2017neural}, with no multiplication and friendly for hardware implementation. Addressing the challenge that deep learning based decoders are difficult to train for long codes \cite{gruber2017deep}, the authors in \cite{cammerer2017scaling} propose to divide the polar coding graph into  sub-blocks, and the decoder for each sub-codeword is trained separately. 

Besides channel decoding, deep learning has the potential to achieve improvement in other areas of communication systems \cite{o2017introduction}. O'Shea \textit{et al.} propose to learn a channel auto-encoder via deep learning technologies \cite{o2016learning}. Farsad \textit{et al.} study the problem of developing detection algorithms for communication systems without proper mathematical models, such as molecular communications \cite{farsad2017detection}. The authors of\cite{samuel2017deep} apply deep neural networks to MIMO detection, which results in comparable detection performance but with much lower complexity. In the application layer, deep recurrent neural networks can be also used to recognize different traffic types \cite{o2016end}. Most recently, Dorner \textit{et al.}\cite{Dorner2017} have demonstrated the feasibility of over-the-air communication with deep neural networks and software-defined radios.

These early studies have not considered the channel decoding problem under correlated Gaussian noise, which is another example of complex channel models that are difficult for theoretical analysis. In this paper, we focus on this problem and propose a novel iterative BP-CNN  receiver for channel decoding.

\section{System design}
\label{sec:system_design}
This section contains the main innovation of this work: an iterative BP-CNN channel decoder. In order to better present the proposed design, we will first describe the system framework. Specifically, the CNN structure will be introduced and its role in helping channel decoding will be explained. Finally, we will highlight two crucial components in training the network:  the design of the loss function and the data generation.


\subsection{System Framework}
\label{sec:system_framework}
\begin{figure}
	\includegraphics[width=\linewidth]{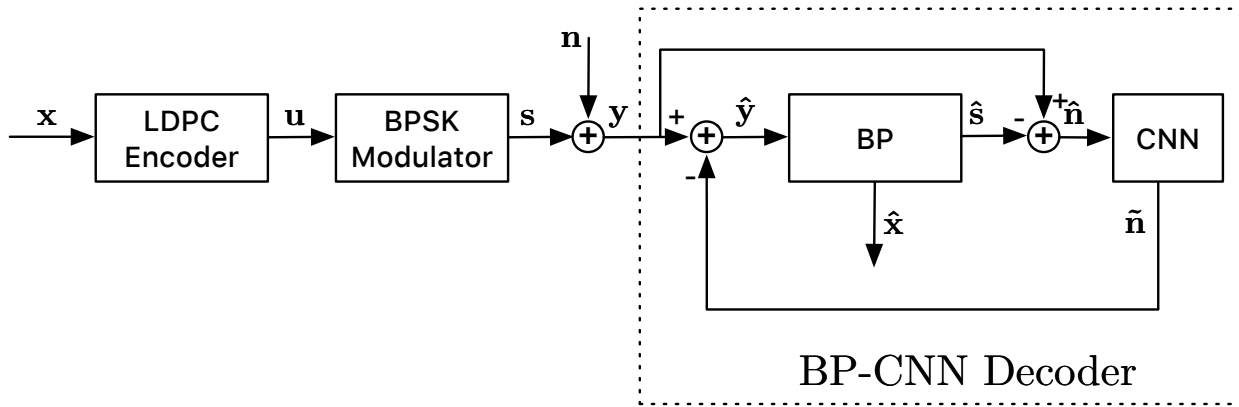}
	\caption{The proposed iterative decoding architecture which consists of a belief propagation (BP) decoder and a feed-forward convolutional neural network (CNN).}
	\label{fig:sys_framework}
\end{figure}
The proposed architecture is illustrated in Fig. \ref{fig:sys_framework}. At the transmitter, a block of uniformly distributed bits $\mathbf{x}$ of length $K$, is encoded to a binary codeword $\mathbf{u}$ of length $N$ through a linear channel encoder. In this paper, we focus on the LDPC code, but the proposed method is readily applicable to other linear block codes. The codeword $\mathbf{u}$ is then mapped to a symbol vector $\mathbf{s}$ through the BPSK modulation. 

The BPSK symbols will be passed through a channel with additive Gaussian noise. The channel noise vector, denoted as $\mathbf{n}$ of length $N$, is modeled as a Gaussian random vector with auto-correlation matrix $\mathbf{\Sigma}$. Note that the LDPC codeword may be long, and as a result the size of $\mathbf{\Sigma}$ can be significant. To better illustrate our design, we will use a standard correlation model that is widely adopted in the literature \cite{sharma2013snr}. The correlation matrix $\mathbf{\Sigma}$ is given by:
\begin{equation}
\label{eqn:corr_model}
\begin{split}
\varSigma_{i,j}=
\begin{cases}
\eta^{j-i},   &i\leq j\\
(\eta^{i-j})^*, &i\geq j,
\end{cases}
\end{split}
\end{equation}
where $\varSigma_{i,j}$ is the $(i,j)$th element of $\mathbf{\Sigma}$, and $\eta$ is the correlation coefficient with $|\eta|\leq 1$. We should emphasize that Eqn.~\eqref{eqn:corr_model} is not fundamental to our proposed design, as it works for any correlation model. This will become evident later and be corroborated in the numerical simulations.

At the receiver, vector $\mathbf{y}$ is received and can be written as
\begin{equation}
\begin{split}
\mathbf{y}=\mathbf{s}+\mathbf{n}.
\end{split}
\end{equation}
A BP decoder is then used to decode the transmitted information bit vector from $\mathbf{y}$. Typically,  there is a BPSK soft demodulator before the BP decoding, to calculate the log-likelihood ratios (LLRs) of the transmitted symbols:
\begin{equation}
\label{eqn:llr_def}
\begin{split}
\ssf{LLR}_{i}^{(1)}=\log\frac{\Pr(\mathbf{s}_i=1|\mathbf{y}_i)}{\Pr(\mathbf{s}_i=-1|\mathbf{y}_i)},
\end{split}
\end{equation}
where $s_i$, $y_i$ denote the $i$th BPSK symbol and the corresponding received  symbol, respectively,  and the superscript $(1)$ to $\ssf{LLR}$ indicates that this is the LLR computation for the initial BP. In our system framework of Fig. \ref{fig:sys_framework}, we omit this module and merge it into the BP decoder for simplicity.

We use $\hat{\mathbf{s}}$ to denote the estimated transmit symbols. Subtracting it from the received symbols $\mathbf{y}$, we obtain $\hat{\mathbf{n}}$ as 
\begin{equation}\label{eqn:est_noise_0_before_cnn}
\begin{split}
\hat{\mathbf{n}} = \mathbf{y} - \hat{\mathbf{s}},
\end{split}
\end{equation}
which can be viewed as an estimation of the channel noise. Because of the decoding errors in the BP decoder, $\hat{\mathbf{n}}$ is not exactly the same as the true channel noise $\mathbf{n}$. We can re-write $\hat{\mathbf{n}}$ as 
\begin{equation}\label{eqn:cnn_input}
\begin{split}
\hat{\mathbf{n}} = \mathbf{n} + \mathbf{\xi},
\end{split}
\end{equation}
where $\mathbf{\xi}$ is the error vector of noise estimation. 

We are now ready to explain the proposed BP-CNN decoding architecture, as shown in the dashed box of Fig.~\ref{fig:sys_framework}. Inspired by the successful application of CNNs in image denoising and super-resolution,  and noticing that the correlation in channel noise $\mathbf{n}$ can be considered as a ``feature'' that may be exploited in channel decoding, we propose to concatenate a CNN after BP, to utilize this correlation to suppress $\mathbf{\xi}$ and get a more accurate estimation of the channel noise. The details of the proposed CNN and the associated training procedure will be presented in subsequent sections. Using $\tilde{\mathbf{n}}$ to denote the CNN output and subtracting it from the received vector $\mathbf{y}$ result in
\begin{equation}
\begin{split}
\hat{\mathbf{y}} &= \mathbf{y} - \tilde{\mathbf{n}},\\
&=\mathbf{s} + \mathbf{n} - \tilde{\mathbf{n}},\\
&=\mathbf{s} + \mathbf{r},
\end{split}
\end{equation}
where $\mathbf{r}=\mathbf{n} - \tilde{\mathbf{n}}$ is defined as the \emph{residual noise}. Then, the new vector $\hat{\mathbf{y}}$ is fed back to the BP decoder and another round of BP decoding is performed. Note that before the second round of BP iterations, the LLR values need to be updated as follows,
\begin{equation}\label{eqn:llr_next_bp}
\begin{split}
\ssf{LLR}_{i}^{(2)}=\log\frac{\Pr(\mathbf{s}_i=1|\mathbf{\hat{y}}_i)}{\Pr(\mathbf{s}_i=-1|\mathbf{\hat{y}}_i)},
\end{split}
\end{equation}
where the superscript $(2)$ to $\ssf{LLR}$ indicates that this is the LLR computation for the subsequent BPs after processing the noise estimate with CNN.  

Obviously, the characteristics of the residual noise $\mathbf{r}$ will affect the calculation in \eqref{eqn:llr_next_bp} and thereby influence the performance of the subsequent BP decoding. Therefore, the CNN should be trained to provide a residual noise that is beneficial for the BP decoding. We propose two strategies for training the CNN and concatenating with BP.
\begin{enumerate}
	\item \emph{Baseline BP-CNN.} The CNN is trained to output a noise estimation which is as close to the true channel noise as possible. This is a standard method which can be implemented using  a quadratic loss function in CNN training. Intuitively, an accurate noise estimation will result in very low power residual noise and thus very little interference to the BP decoding. In order to calculate the LLR as \eqref{eqn:llr_next_bp}, we also need to obtain an empirical probability distribution of the residual noise.
	\item \emph{Enhanced BP-CNN.} The baseline BP-CNN may not be optimal considering that the channel encoding (such as LDPC) is mostly optimized for the AWGN channel and the residual noise may not necessarily follow a Gaussian distribution. Therefore, another strategy is to depress the residual noise power and simultaneously let the residual noise follow a Gaussian distribution as much as possible. By doing this, the calculation in \eqref{eqn:llr_next_bp} becomes much easier:
	\begin{equation}\label{eqn:llr_in_awgn}
	\begin{split}
	\ssf{LLR}^{(2)}=\frac{2\hat{y}}{\sigma_r^2},
	\end{split}
	\end{equation}
	where $\sigma_r^2$ is the power of the residual noise.
\end{enumerate}

In Section \ref{sec:training}, we will show how to train the network with the above two strategies. With a depressed influence of noise, the BP decoding is expected to yield a more accurate result. The above processes can be executed iteratively to successively depress the noise influence and improve the final decoding performance.


\begin{remark}
Although majority of the channel coding study focuses on the AWGN or fading channels with i.i.d. noise, we note that  channels with additive correlated noise are particularly important for some systems, such as in wireless and cellular communications. In these systems, coherent fading either in time or frequency combined with multi-user interference can lead to the \textit{equivalent} channel noise being correlated for the purpose of channel decoding. These are increasingly important scenarios for the performance of channel coding, with the growing emphasis on delay-sensitive and mission-critical applications in 5G where long codes and deep interleaving may not be appropriate \cite{Liva2016}. 

%

\end{remark}

\begin{remark}
We note that the CNN in our system is a pure feed-forward network without memory and state information. In the deep learning literature, there is another celebrated network form called \textit{recurrent neural networks} (RNN), which is mainly used in language processing \cite{mikolov2010recurrent}. The exploitation of the RNN structure with memory in BP decoding is an important and interesting topic that we plan to investigate in the  future.
\end{remark}

\subsection{Why  is CNN Useful for Channel Decoding?}
As we have mentioned, one of the first motivations for us to use a CNN to estimate the channel noise comes from CNN's successful applications in some low-level tasks such as image denoising \cite{jain2009natural,zhang2016beyond} and image super-resolution \cite{dong2016image}. A careful investigation into these CNN applications reveals important similarity to the channel decoding task at hand. Let us take image denoising as an example. Note that the task of image denoising is to recover the original image pixels $\mathbf{X}$ from its noisy observation $\mathbf{Y}$, following the additive model $\mathbf{Y}=\mathbf{X}+\mathbf{W}$ where $\mathbf{W}$ denotes the unknown noise. As analyzed in \cite{jain2009natural}, there is a strong mathematical relationship between convolution networks and the state-of-the-art Markov random field (MRF) methods. One advantage to utilize a CNN for image denoising is that it allows the network to learn high dimensional model parameters to extract image features, which have stronger image processing power. The results in \cite{jain2009natural} demonstrate that using CNN can achieve a better performance with low complexity than previously reported methods.


Returning to our task in this paper, the role of CNN in the decoding architecture is to recover the true channel noise $\mathbf{n}$ from the noisy version $\hat{\mathbf{n}}$ following model \eqref{eqn:cnn_input}, which is very similar to image denoising. The recovered channel noise $\mathbf{n}$ in our task corresponds to the image pixels $\mathbf{X}$ in image denoising and the error $\mathbf{\xi}$ corresponds to the noise $\mathbf{W}$. Note that just like CNN-based image denoising which exploits the image features in pixels $\mathbf{X}$, the iterative BP-CNN decoder exploits the  correlations in channel noise $\mathbf{n}$. Given that CNN has excellent denoising performance with low complexity, it is natural to ask if such advantage can translate to the channel decoding task in this paper.


\subsection{Belief Propagation Decoding}
\label{sec:BP_decoder}
In order to make this paper self-contained, we briefly introduce the standard belief propagation (BP) decoding process. BP is an iterative process in which messages are passed between the variable nodes and check nodes in a Tanner graph.  We use $v$ to denote a variable node and $c$ for a check node. $L_{v\rightarrow c}$ and $L_{c\rightarrow v}$ are messages passed from $v$ to $c$ and $c$ to $v$, respectively. First, $L_{v\rightarrow c}$ is initialized with the LLR value calculated from the received symbol, as shown in (\ref{eqn:llr_def}). Then the messages are updated iteratively as follows
\begin{equation}
\begin{split}
L_{v\rightarrow c}&=\sum_{c'\in \mathcal{N}(v)\backslash c} L_{c'\rightarrow v}, \\
L_{c\rightarrow v}&=2\tanh^{-1}\left(\prod_{v'\in \mathcal{N}(c)\backslash v}\tanh\frac{L_{v'\rightarrow c}}{2}\right),
\end{split}
\end{equation}
where $\mathcal{N}(v)\backslash c$ $\left(\mathcal{N}(c)\backslash v\right)$ denotes the set of neighboring check (variable) nodes of the variable node $v$ (the check node $c$), except $c$ ($v$). After several iterations, the LLR of a variable node $v$, denoted as $L_v$, is calculated as
\begin{equation}
\begin{split}
L_v=\sum_{c'\in \mathcal{N}(v)} L_{c'\rightarrow v}.
\end{split}
\end{equation}
Finally, the bit value corresponding to the variable node $v$ is determined by
\begin{equation}
b=
\begin{cases}
0, & \text{  if } L_v\geq 0, \\
1, & \text{  if } L_v\le 0.
\end{cases}
\end{equation}

The standard BP decoding is well known to be effective in AWGN channels \cite{LC:04} but will have difficulty to handle correlation in the channel noise. To see this point, we recall that the BP decoding algorithm operates on a factor graph with the following probability model for the AWGN channel:
\begin{equation}\label{eqn:factor_model_standard_bp}
\begin{split}
p(x_1,x_2,...,x_N)=\left(\prod_i p_{ch}(x_i)\right)\left(\prod_i f_i(\mathcal{N}_i)\right),
\end{split}
\end{equation} 
where $x_i$ is the $i$th bit, $p(x_1,x_2,...,x_N)$ denotes the joint probability of all bits, $p_{ch}(x_i)$ is the probability conditioned on the received channel symbols, $f_i(\cdot)$ denotes the $i$th parity-check indicator function corresponding to the $i$th check node, and $\mathcal{N}_i$ denotes the set of variable nodes connected to the $i$th check node. For a decoder facing correlated noise, however, the probability model can only be presented in the following form:
\begin{equation}\label{eqn:factor_model_colored_channel}
\begin{split}
p(x_1,x_2,...,x_N)=p_{ch}(x_1, x_2,...,x_N)\left(\prod_i f_i(\mathcal{N}_i)\right),
\end{split}
\end{equation} 
where the probability conditioned on the received symbols cannot be factored as symbols are correlated. Therefore, improving the standard BP decoding algorithm to incorporate the joint  distribution of noise samples may have very high complexity, especially when the channel noise has strong correlations. 



\subsection{CNN for  Noise Estimation}
As mentioned before, adopting CNN for noise estimation is enlightened by its successful applications in computer vision, which demonstrate its strong capability to extract local features. For some specific image restoration tasks, Dong \textit{et al.} \cite{dong2016image} have shown that CNN has a similar performance as previously known strategies which represent image patches by a set of pre-trained bases, but with a lower complexity. Moreover, the network can be trained to find better bases instead of the  pre-defined ones, for better performance. Therefore,  CNN has the potential to yield a better image restoration quality. 

\begin{figure*}
	\includegraphics[width=\linewidth]{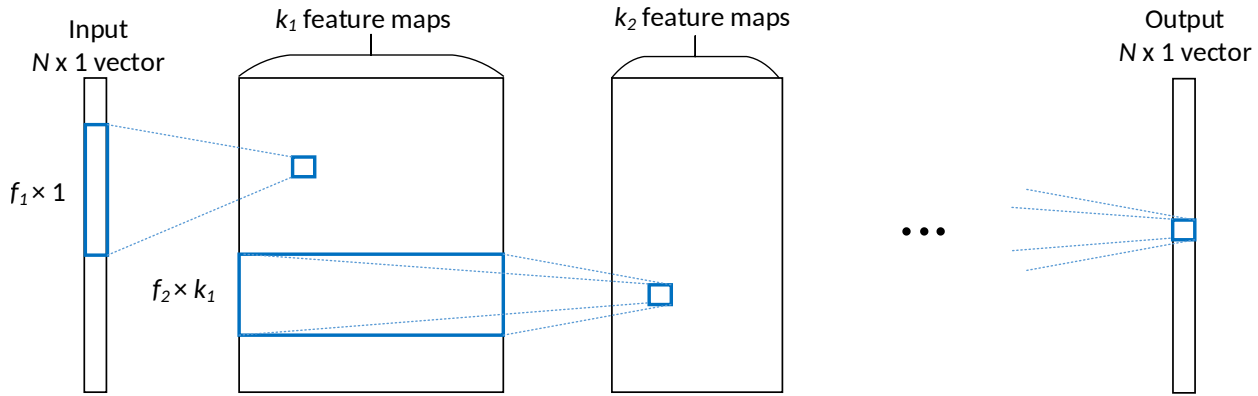}
	\caption{The adopted CNN structure for noise estimation.}
	\label{fig:cnn_structure}
\end{figure*}

In the proposed iterative BP-CNN architecture, the adopted network structure, which is shown in Fig. \ref{fig:cnn_structure},  is similar to those used for low-level tasks in image restoration \cite{jain2009natural,dong2016image} but with a conspicuous difference in that the input of our network is a 1-D vector instead of a 2-D image. As can be seen at the first layer in Fig. \ref{fig:cnn_structure}, $k_1$ feature maps are generated from the input data $\hat{\mathbf{n}}$, which can be expressed as
\begin{equation}\label{eqn:conv_layer}
\begin{split}
\mathbf{c}_{1,j} = \ssf{ReLU}(\mathbf{h}_{1,j}*\hat{\mathbf{n}} + b_{1,j}),
\end{split}
\end{equation}
where $\mathbf{c}_{1,j}$ is the $j$th feature map at the first layer, $\mathbf{h}_{1,j}$ is the $j$th convolution kernel, which is essentially a 1-D vector of length $f_j$, $b_{1,j}$ is the corresponding bias, and $\ssf{ReLU}(\cdot)$ is the Rectified Linear Unit function ($\max(x,0)$) in order to introduce nonlinearity \cite{nair2010rectified}. In neural networks, the convolution operation is denoted by $*$ in \eqref{eqn:conv_layer} and is defined as 
\begin{equation}\label{eqn:conv_ops}
\begin{split}
\left(\mathbf{h}_{1,j}*\hat{\mathbf{n}}\right)(v)=\sum_{\Delta v}\hat{\mathbf{n}}(v+\Delta v)\mathbf{h}_{1,j}(\Delta v),
\end{split}
\end{equation}
which is slightly different with the standard definition in signal processing. 

At the $i$th layer ($i>1$), the convolution operation is performed over all feature maps of the previous layer, which can be viewed as a 2-D convolution as shown in Fig. \ref{fig:cnn_structure}. The output can be written as
\begin{equation}
\begin{split}
\mathbf{c}_{i,j} = \ssf{ReLU}(\mathbf{h}_{i,j}*\mathbf{c}_{i-1} + b_{i,j}),
\end{split}
\end{equation}
where $\mathbf{c}_{i,j}$ is the $j$th feature map at the $i$th layer. $\mathbf{h}_{i,j}$ is the $j$th convolution kernel of size $f_i\times k_{i-1}$ where $k_{i-1}$ is the number of feature maps of the previous layer.
We use $L$ to denote the number of total layers. At the last layer, the final estimation of the channel noise is
\begin{equation}
\begin{split}
\tilde{\mathbf{n}}=\mathbf{h}_{L}*\mathbf{c}_{L-1} + b_{L}.
\end{split}
\end{equation}

To summarize, the structure of a CNN is determined by its number of layers, filter sizes, and feature map numbers in each layer. These parameters need to be decided before training the network. For simplicity of exposition, we denote the structure of the network as
\begin{equation}\label{eqn:cnn_structure}
\{L;f_1, f_2,..,f_L; k_1, k_2,...,k_L\}.
\end{equation}

Besides convolution layers, \textit{pooling}, \textit{dropout} and \textit{fully-connected} layers are also important components in ordinary CNN architectures. However, they are not considered in our design for the following reasons. First, pooling layers are often used for downsampling in  high-level tasks. In our problem, the CNN output is still a low-level representation which has the same dimension as the CNN input. Pooling layers are not needed in this case, because it may discard some useful details \cite{mao2016image}. Second, dropout is a widely used technique to control overfitting. In this work, we already have overfitting controlled by tracking the loss value in the validation set. Furthermore, we have tested the performance of dropout and found that it does not provide any gain. With regard to the fully connected layers, they are often used to output the high-level description after the data dimensions have been reduced to a large extent. In the low-level tasks, fully-connected layers are of high complexity and also hard to train. 

\subsection{Training}
\label{sec:training}
\subsubsection{Loss function}
It is well known in deep learning that the performance of a network depends critically on the choice of loss functions for training \cite{Goodfellow2016}. Generally speaking, a loss function is to measure the difference between the actual CNN output and its expected output, and it should be carefully defined based on the specific task of the network. In the proposed architecture, CNN is used to estimate the channel noise and its output will influence the performance of BP decoding in the next iteration. Therefore, a proper loss function must be selected by fully considering the relationship between CNN and the following BP decoding.

As introduced in Section \ref{sec:system_framework}, to facilitate the subsequent BP decoding, there are two strategies to train the network. One is to only depress the residual noise power (\emph{baseline BP-CNN}), and the other one is to depress the residual noise power and simultaneously shape its distribution (\emph{enhanced BP-CNN}).

For \emph{baseline BP-CNN}, the loss function can be chosen as the typical residual noise power: 
\begin{equation}
\begin{split}
\ssf{Loss}_{\text{A}}=\frac{\|\mathbf{r}\|^2}{N},
\end{split}
\label{eqn:squraed_loss}
\end{equation}
where $N$ is the length of a coding block. Note that this is the well adopted quadratic cost function in neural network training.

For \emph{enhanced BP-CNN}, how to define a proper loss function is an open problem. well-known loss functions such as quadratic, cross-entropy, or Kullback-Leibler divergence does not accomplish the goal. In this work, we introduce \textit{normality test} to the loss function, so that we can measure how likely the residual noise samples follow a Gaussian distribution. The new loss function is formally defined as
\begin{equation}
\label{eqn:loss_func}
\begin{split}
\ssf{Loss}_{\text{B}} = \frac{\|\mathbf{r}\|^2}{N} + \lambda \left(S^2+\frac{1}{4}\left(C-3\right)^2\right).
\end{split}
\end{equation}
The first term in (\ref{eqn:loss_func}) measures the power of the residual noise and the second term, adopted from the Jarque-Bera test \cite{thadewald2007jarque}, represents a normality test to determine how much a data set is modeled by a Gaussian distribution. $\lambda$ is a scaling factor that balances these two objectives. Specifically, $S$ and $C$ in \eqref{eqn:loss_func} are defined as follows:
\begin{equation}
\label{eqn:jb_test}
\begin{split}
S&= \frac{\frac{1}{N}\sum_{i=1}^{N}\left(r_i-\bar{r}\right)^3}{\left(\frac{1}{n}\sum_{i=1}^{N}\left(r_i-\bar{r}\right)^2\right)^{3/2}},\\
C&= \frac{\frac{1}{N}\sum_{i=1}^{N}\left(r_i-\bar{r}\right)^4}{\left(\frac{1}{n}\sum_{i=1}^{N}\left(r_i-\bar{r}\right)^2\right)^{2}},
\end{split}
\end{equation}
where $r_i$ denotes the $i$th element in the residual noise vector, and $\bar{r}$ is the sample mean. In statistics, $S$ and $C$ are called \textit{skewness} and \textit{kurtosis}, respectively. Although the Jarque-Bera test is not the optimal normality test, the cost function is derivable and simple for training. Furthermore,  experimental results show that it can provide a very desirable output. 

\subsubsection{Generating the training data}
To train the network, we need both channel noise data $\mathbf{n}$ and estimated noise data $\hat{\mathbf{n}}$ from the BP decoding results. For the simulation results in Section~\ref{sec:experiment}, we focus on the channel models with known (to the receiver) noise correlation functions, and thus we are able to generate adequate channel noise samples to train the network. For example, given the channel correlation matrix $\mathbf{\Sigma}$ in \eqref{eqn:corr_model}, the channel noise samples can be generated by
\begin{equation}
\begin{split}
\mathbf{n}=\mathbf{\Sigma}^{1/2}\mathbf{n}_w,
\end{split}
\end{equation}
where $\mathbf{n}_w$ is an AWGN noise vector. The estimated noise $\hat{\mathbf{n}}$ can be obtained by generating uniform distributed binary bits $\mathbf{x}$ and successively performing channel encoding, BPSK mapping, simulated channel interference and BP decoding. 

Another factor in generating the training data is the channel condition, i.e., the signal to noise ratio (SNR), which will determine the severity of errors of the BP decoding and hence affect the input to the network. We use $\Gamma$ to denote a set of channel conditions to generate training data. If the channel condition is very good, very few errors exist $\hat{\mathbf{n}}$ and the network may not learn the robust features of the channel noise. On the other hand, if the channel condition is very bad, many errors exist in  $\hat{\mathbf{n}}$  and they will mask the channel noise features, which is also detrimental for the network training.  In Section \ref{sec:experiment}, we will conduct some experiments to analyze this problem.

We want to emphasize that although CNN training typically requires a large amount of data, has high complexity, and the resulting network depends on the training data, these factors will not impede the application of the proposed architecture to practical systems. The network training is largely done offline, where the required training complexity can be satisfied with powerful computational devices such as GPUs. The amount of training data is also an offline issue, which can be generated and stored in mass storage. Lastly, although CNN depends on the training data, in practical communication system design we often focus on certain representative scenarios (``use cases'') and we can generate the training data to reflect these scenarios.  The resulting trained CNNs can be stored in the on-device memory for online use.

\subsection{Design Summary}
\label{sec:system_summary}

\begin{table}
	\caption{Summary of important system parameters.}
	\centering
	\begin{tabular}{|l|l|}
		\hline 
		\emph{Notation} & \emph{Meaning}  \\ 
		\hline
		$K$ & \tabincell{l}{The number of iterations between\\ the BP decoder and the CNN} \\
		\hline
		$\{L;f_1, ...,f_L; k_1, ...,k_L\}$ & \tabincell{l}{CNN structure: \\ numbers of layers, filter sizes \\and numbers of feature maps} \\
		\hline
		$\{B\}$ & Numbers of BP iterations \\
		\hline
		$\lambda$ & \tabincell{l}{The scaling factor of the \\normality test for \emph{enhanced BP-CNN}} \\
		\hline
		$\Gamma$ & \tabincell{l}{Channel conditions to generate \\training data}\\
		\hline 
	\end{tabular}
	\label{tab:sys_parameters}
\end{table} 

The proposed design can be summarized using the relevant parameters, which will result in the tradeoff between performance and complexity as we will see later in the simulations.  First, noting that the proposed decoding framework is an iterative architecture, one important parameter is the number of total iterations, denoted as $K$. Then, the structure of CNN is decided by the number of layers, filter sizes and the number of feature maps in each layer, denoted as $\{L;f_1, f_2,...,f_L; k_1, k_2,...,k_L\}$. For the BP decoder, the  parameter of interest is the number of iterations. Essentially, the iterative BP-CNN decoding architecture in Fig. \ref{fig:sys_framework} can be unfolded to generate an open-loop structure and all networks are trained sequentially, which can be concisely denoted as BP-CNN1-BP-CNN2-...-CNN$x$-BP. The proposed iterative architecture is actually a subset of this general framework.  Note that although the open-loop framework is more general and thus may result in better performance, it will consume a much larger amount of resource to implement. The closed-loop architecture in Fig. \ref{fig:sys_framework} only requires training and storing one CNN, while the open-loop framework has $x$ CNNs. The training of multiple serially-concatenated CNNs may be highly complex, as a later CNN may depend on the training of an early one. Storage of $x$ CNNs on the device will also increase the cost.  

To train a network with \emph{enhanced BP-CNN}, we need to select a proper $\lambda$ to balance the significance of normality test in the loss function and the channel conditions to generate training data. Important system parameters that need to be carefully chosen are summarized in Table \ref{tab:sys_parameters}.  In Section \ref{sec:experiment}, we will carry out extensive experiments to analyze the influence of these system parameters on the performance and provide some guidance on their selection.

\section{Experiments}
\label{sec:experiment}
\subsection{Overview}
\begin{table}
\caption{Basic CNN setting for experiments.}
\centering
\begin{tabular}{|c|c|}
	\hline 
	CNN structure & $\{4;9,3,3,15;64,32,16,1\}$  \\ 
	\hline 
	Mini-batch size & 1400  \\ 
	\hline 
	Size of the training data & 2000000 \\
	\hline
	Size of the validation data & 100000 \\
	\hline
	\tabincell{c}{SNRs for generating \\the training data ($\Gamma$)} & $\{0,0.5,1,1.5,2,2.5,3\}$ dB\\
	\hline
	Initialization method & Xavier initialization \\
	\hline
	Optimization method & Adam optimization \\
	\hline
\end{tabular}
\label{tab:basic_cnn_settings}
\end{table} 

Throughout all experiments, we use a LDPC code of rate $3/4$. The code block length is 576 and the parity check matrix is from \cite{helmling2016database}. The experimental platform is implemented in TensorFlow \cite{abadi2016tensorflow}. The basic CNN structure is $\{4;9,3,3,15;64,32,16,1\}$. The network is initialized using the Xavier initialization method, which has been demonstrated to perform better than random initialization \cite{glorot2010understanding}. We use the Adam optimization method for searching the optimal network parameters \cite{kingma2014adam}.  During the training process, we check the loss value over the validation set every 500 iterations. The training data is generated under multiple channel SNRs: $\{0,0.5,1,1.5,2,2.5,3\}$ dB. The channel SNR is defined as
\begin{equation}\label{eqn:snr}
\ssf{SNR}=10\log\left(\frac{P}{\sigma^2}\right),
\end{equation}
where $P$ is the average transmit power and $\sigma^2$ is the average power of the complex noise samples in the I-Q plane. Each mini-batch contains 1400 blocks of data and the data of each SNR occupies the same proportion. Training continues until the loss does not drop for a consecutive period of time (eight checks in our experiment, which equals to 4000 iterations). The basic CNN setting is summarized in Table \ref{tab:basic_cnn_settings}. 

In all experiments, the setting is the same as Table \ref{tab:basic_cnn_settings} unless otherwise specified. The system performance is measured by the bit error rate (BER). In order to test BERs with high accuracy and low complexity, a total of $10^8$ information bits are tested for high SNRs and $10^7$ bits for low SNRs. 

In the remainder of this section, we will analyze and verify the proposed iterative BP-CNN decoding architecture from different perspectives.

\subsection{Performance Evaluation}
\subsubsection{BP-CNN reduces decoding BER}

\begin{figure}
	\centering
	\subfigure[$\eta=0.8$, strong correlation. $\lambda=0.1$.]{ \includegraphics[width=0.5\linewidth]{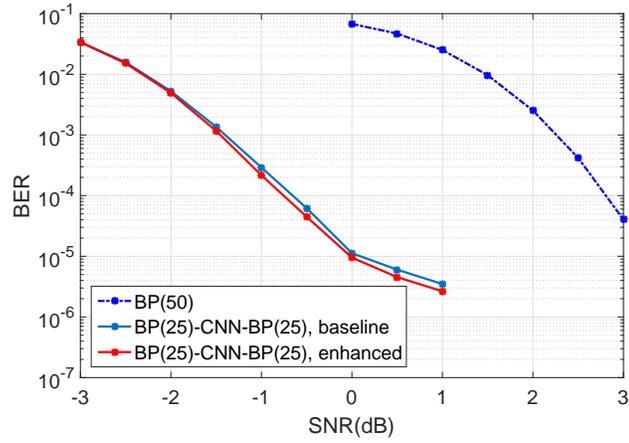}}
	\subfigure[$\eta=0.5$, moderate correlation. $\lambda=10$.]{ \includegraphics[width=0.5\linewidth]{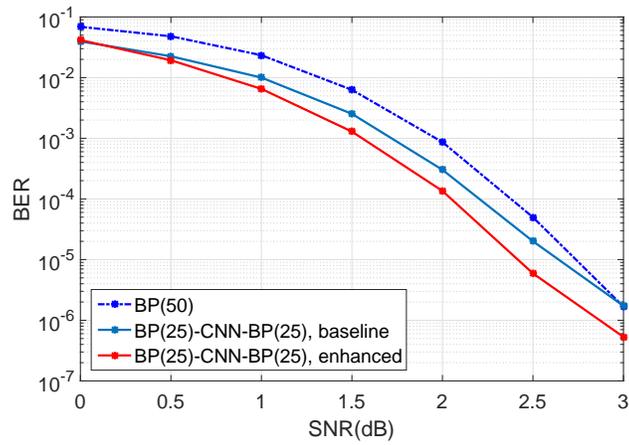}}
	\subfigure[$\eta=0$, no correlation. $\lambda=10$.]{ \includegraphics[width=0.5\linewidth]{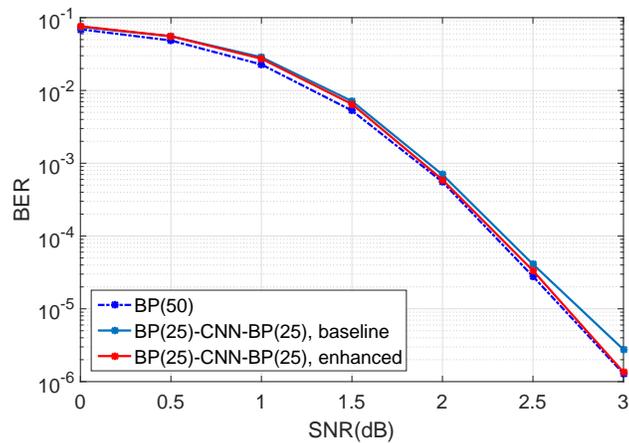}}
	\caption{Performance comparison of BP-CNN with the standard BP decoding. Only one iteration between CNN and BP decoder is executed for the proposed design. The numbers in the brackets denote the BP iterations.}
	\label{fig:exp_syscomp_one_iteration}
\end{figure}

We first compare the performances of the proposed method and the standard BP decoder. For the BP decoder, total 50 iterations are executed (denoted as ``BP(50)'' in the figure). For the proposed method, we use its simplest form for testing, i.e. there is only one iteration between the BP decoder and the CNN. In this case, the receiver structure can be simply denoted as BP-CNN-BP. We present the comparison results for two correlation parameters: $\eta=0.8$ represents a relatively strong correlation model and $\eta=0.5$ represents a moderate one. In addition, the test results under an AWGN channel without any correlation ($\eta=0$) are presented to demonstrate that the proposed method can also work with uncorrelated noises, and thus has a wide range of application. Both \emph{baseline} and \emph{enhanced} BP-CNNs are tested. For \emph{enhanced BP-CNN}, we set $\lambda$ to 0.1, 10 and 10 for $\eta=0.8$, $0.5$ and $0$, respectively. 

In order to isolate and identify the contribution of CNN, we have kept the total number of BP iterations the same in two systems. In the BP-CNN method, we execute 25 BP iterations in each BP decoding process (denoted as ``BP(25)-CNN-BP(25)'' in the figure), resulting in the same 50 BP iterations as the standard BP decoder. 

\begin{figure}
	\centering
	 \includegraphics[width=0.8\linewidth]{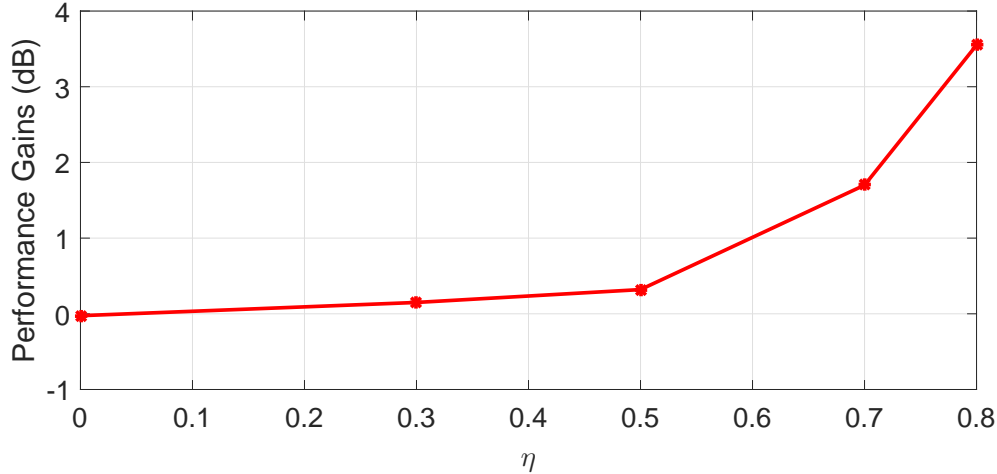}
	\caption{Performance gains of  \emph{enhanced BP-CNN} under different $\eta$'s. $\lambda = 10$ for $\eta=0,0.3,0.5$ and $\lambda= 0.1$ for $\eta=0.7,0.8$.}
	\label{fig:gain_vs_eta}
\end{figure}

The experiment results are reported in Fig. \ref{fig:exp_syscomp_one_iteration}. We can see that both baseline and enhanced BP-CNN achieve significant performance gains with correlated noise. In the strong correlation case when $\eta=0.8$, BP-CNN can improve the decoding performance by approximately 3.5dB at BER=$10^{-4}$. It should be emphasized that this performance gain cannot be compensated by more iterations in the standard BP decoder, as BP(50) already achieves a saturating performance. In the moderate correlation case with $\eta=0.5$, the performance gain becomes smaller, because the correlation is weaker and the benefit of adopting a CNN is less. For the special case where $\eta=0$ and the noise becomes i.i.d., which is the nominal AWGN channel, the proposed method performs similarly with the standard BP decoding. We thus conclude that the iterative BP-CNN decoding method can support a wide range of correlation levels, and the performance gains vary adaptively with the noise correlation. To see this more clearly, we plot the SNR gains at BER=$10^{-4}$ of the enhanced BP-CNN decoder over the standard BP decoding algorithm under different values of $\eta$ in Fig.~\ref{fig:gain_vs_eta}. We see that the performance gain grows \textit{monotonically} with the correlation level $\eta$, which is as expected because higher $\eta$ presents a better opportunity for CNN to extract the noise ``feature''.

We can also compare the baseline and enhanced BP-CNN decoders from  Fig. \ref{fig:exp_syscomp_one_iteration}. We see that although both methods outperform the standard BP, \emph{enhanced BP-CNN} further outperforms the baseline strategy. As explained in Section~\ref{sec:training}, the enhanced strategy balanced both accurate noise estimation and shaping the output empirical distribution, and thus is better suited for concatenation with the BP decoder.


\subsubsection{BP-CNN achieves performance gain with lower complexity}

\begin{figure}
	\centering
	\subfigure[$\eta=0.8$, strong correlation. $\lambda=0.1$.]{ \includegraphics[width=0.6\linewidth]{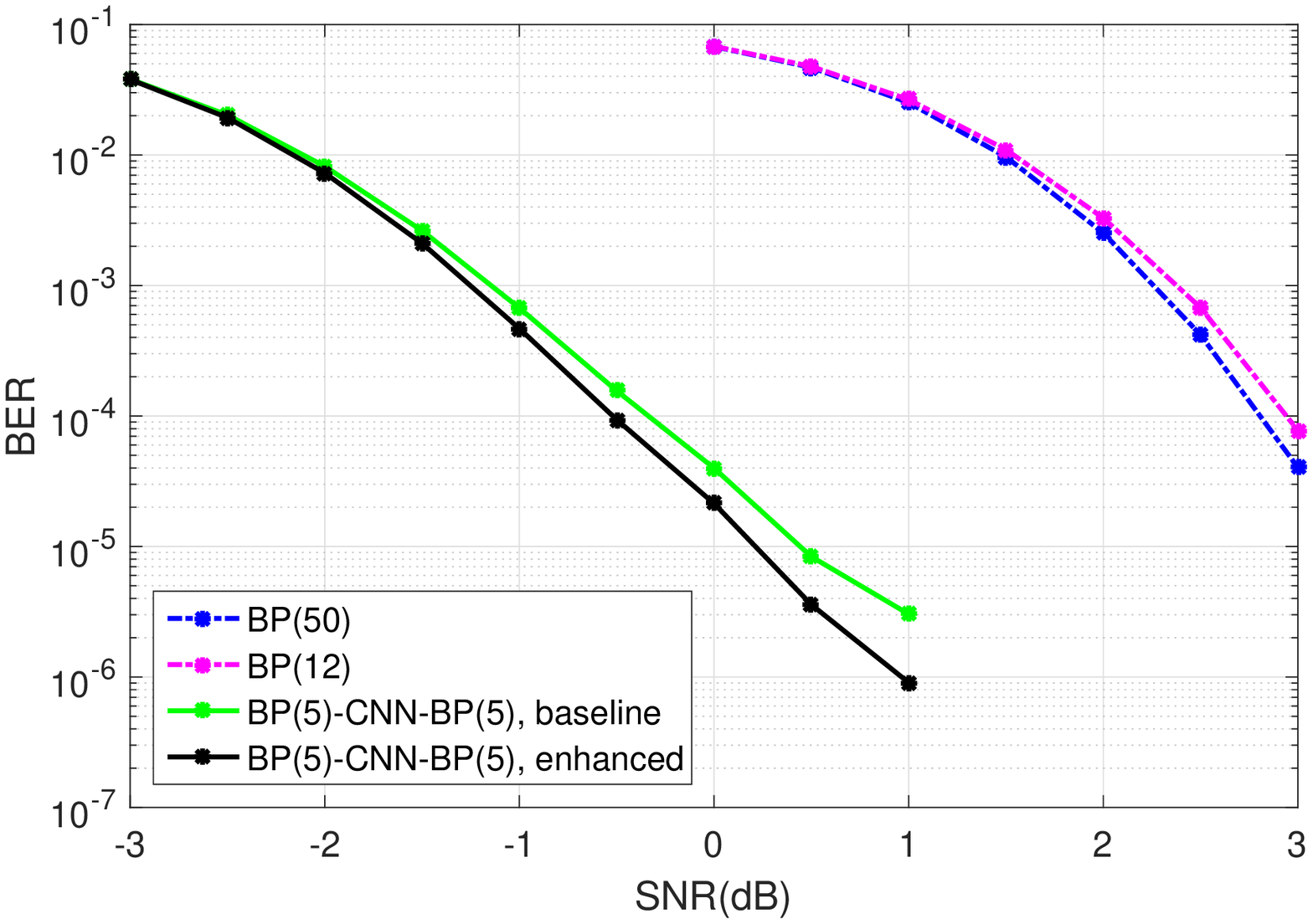}\label{fig:exp_syscomp_lowcomplexitya}}
	\subfigure[$\eta=0.5$, moderate correlation. $\lambda=10$.]{ \includegraphics[width=0.6\linewidth]{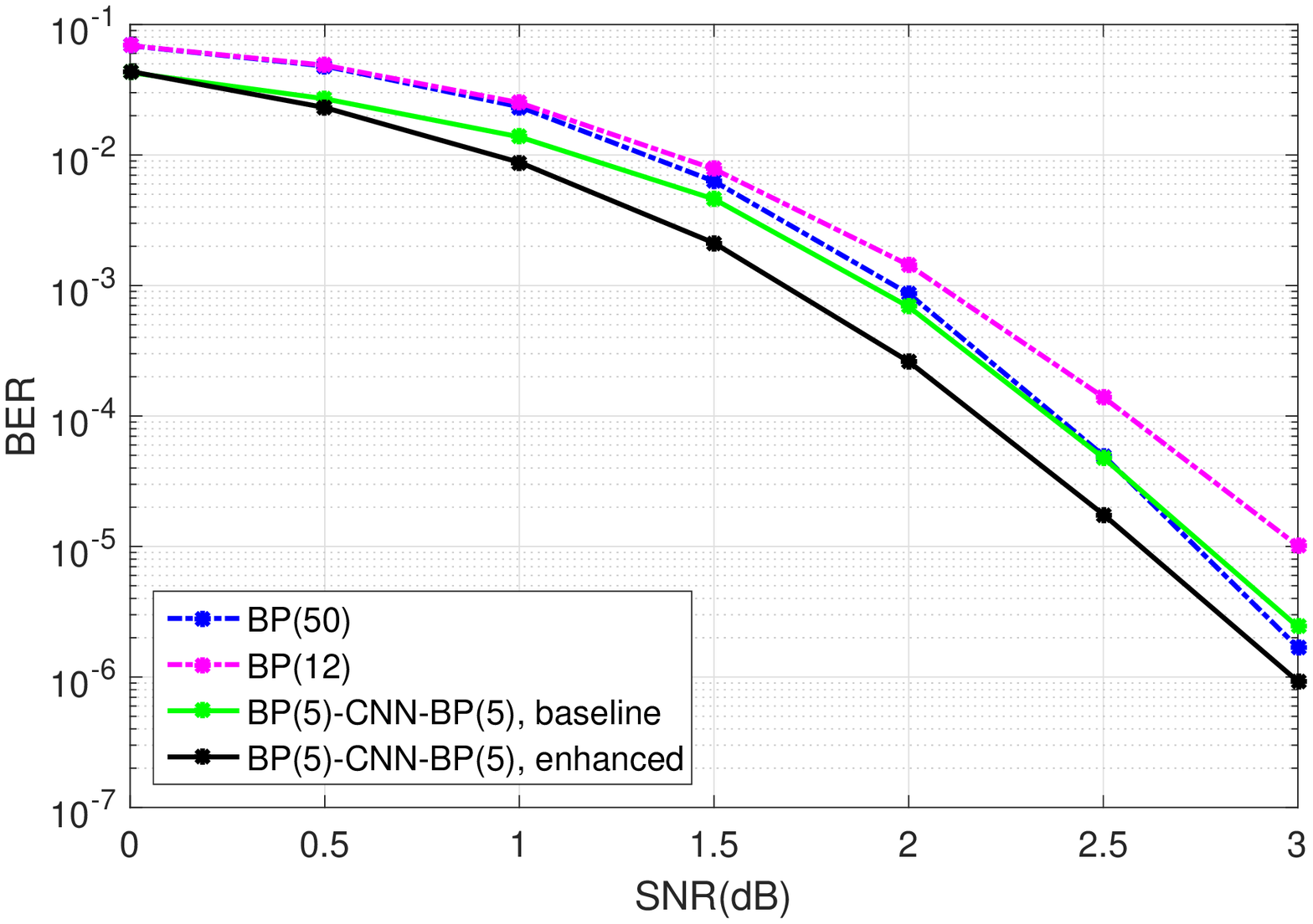}\label{fig:exp_syscomp_lowcomplexityb}}
	\caption{BP-CNN achieves performance gains with lower complexity.}
	\label{fig:exp_syscomp_lowcomplexity}
\end{figure}

Besides improving the decoding performance under the same number of BP iterations, another desirable feature of the iterative BP-CNN decoder is that it can outperform the standard BP decoding with lower overall complexity. To see this, we present another set of results in Fig. \ref{fig:exp_syscomp_lowcomplexity}. The implementation details are the same as Fig. \ref{fig:exp_syscomp_one_iteration} except that in the BP-CNN decoder, the number of each BP decreases from 25 to 5. For comparison, we also plot the standard BP decoding performance with 12 and 50 iterations, respectively. Ideally, we would like to set the BP parameters so that both methods have exactly the same overall complexity, and then compare their BER performances. However, an accurate comparison of the complexity associated with a BP network and a CNN is quite difficult. We get around this problem by comparing their runtime under the same computation environment\footnote{Both are implemented in TensorFlow and simulations are run with the same computation resources.} and adjusting the parameters accordingly.  In our test environment, we observe that the runtime of CNN with the structure $\{4;9,3,3,15;64,32,16,1\}$ is roughly equivalent to two BP iterations. This means that the chosen BP(5)-CNN-BP(5) structure in Fig. \ref{fig:exp_syscomp_lowcomplexity} has approximately the same complexity as the standard BP decoder with 12 iterations. We also plot BP(50) and observe that further increasing the complexity of standard BP provides some marginal gain. Hence,  BP(12) represents a good-but-not-saturated scenario where the comparison to BP(5)-CNN-BP(5) is fair and meaningful.

We see from Fig.~\ref{fig:exp_syscomp_lowcomplexity} that the baseline BP(5)-CNN-BP(5) decoder has comparable performance of standard BP(50), but with a much lower complexity. When comparing the decoding performance with approximately the same complexity, both baseline and enhanced BP(5)-CNN-BP(5) decoders outperform the standard BP(12) decoder. Again, this gain is significant (around 3dB) in the strong correlation case of Fig.~\ref{fig:exp_syscomp_lowcomplexitya} but still noticeable (0.1 to 0.5dB) in the moderate correlation case of Fig.~\ref{fig:exp_syscomp_lowcomplexityb}.

As the {enhanced} BP-CNN decoder outperforms the {baseline} one, we focus on the enhanced BP-CNN in the remainder of the experiments.


\subsubsection{Choice of hyperparameter $\lambda$ affects the performance of enhanced BP-CNN}

\begin{figure}
	\centering
	\subfigure[$\eta=0.8$, strong correlation.]{ \includegraphics[width=0.6\linewidth]{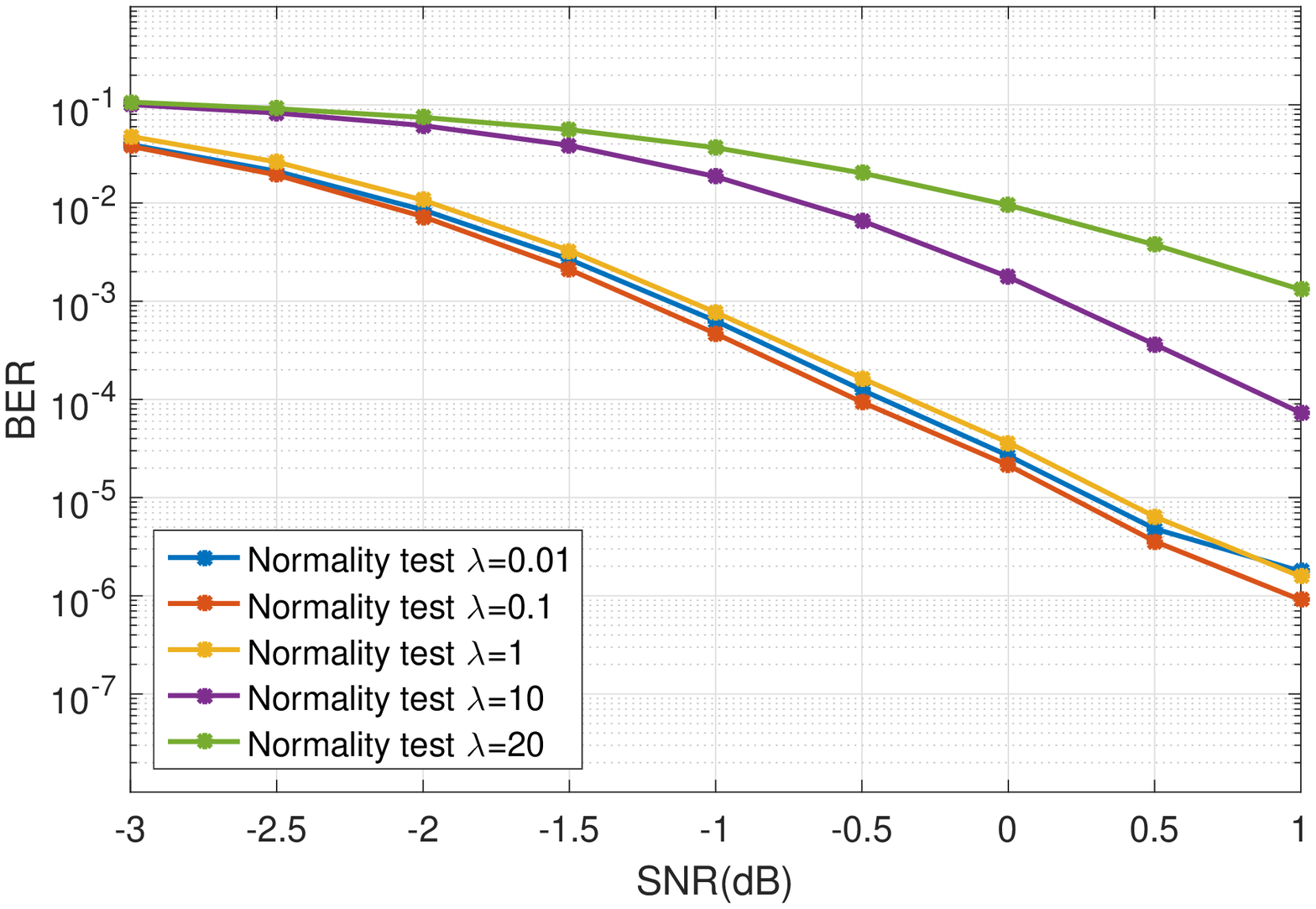}}
	\subfigure[$\eta=0.5$, moderate correlation.]{ \includegraphics[width=0.6\linewidth]{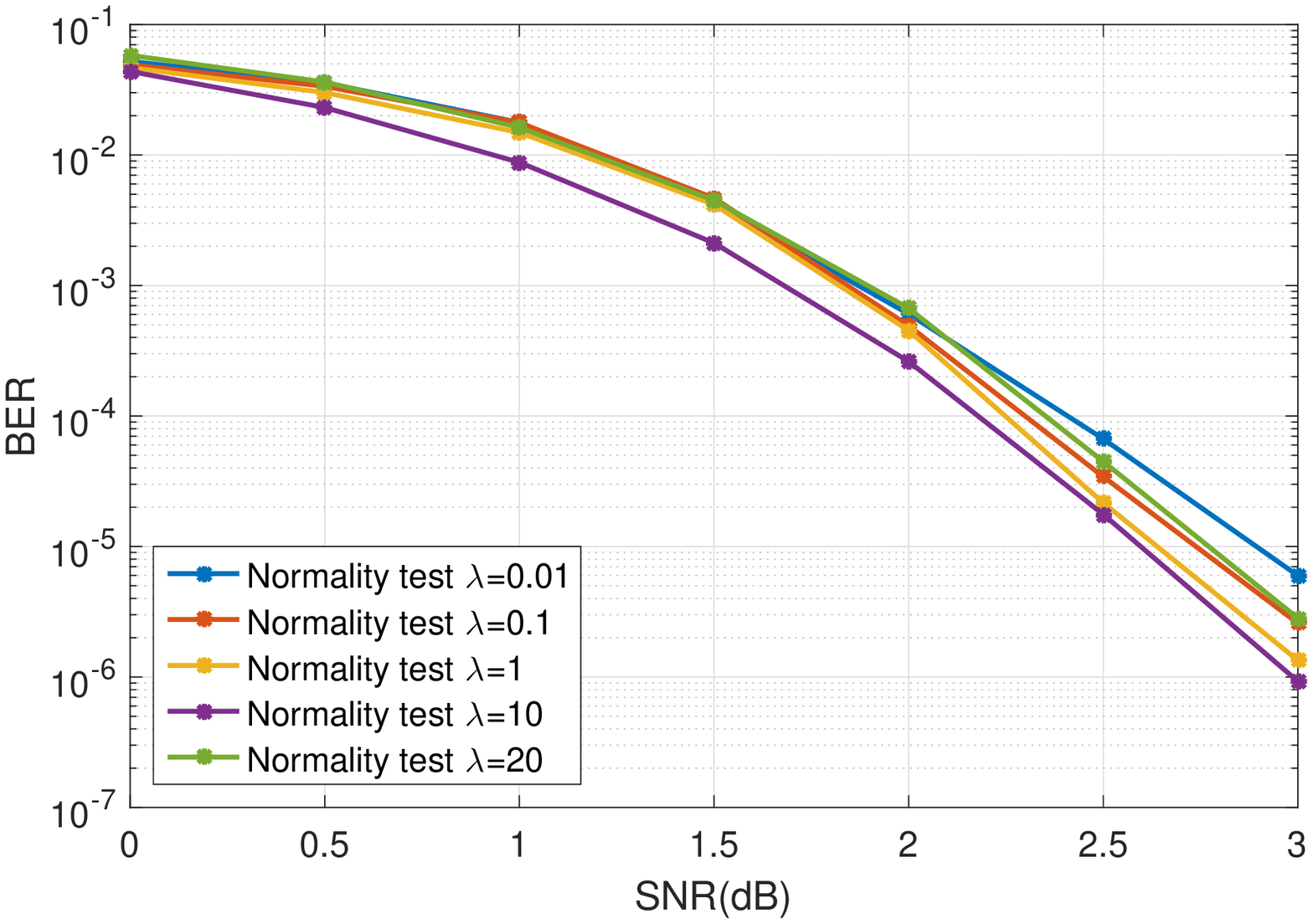}}
	\caption{Showing the necessity of the normality test. The receiver structure is BP(5)-CNN-BP(5).}
	\label{fig:necessity_norm_test}
\end{figure}

In \eqref{eqn:loss_func}, we have defined a new loss function for {enhanced BP-CNN}, which involves measures of both \emph{power} and  \emph{normality} of the residual noise. The coefficient $\lambda$ is an important hyperparameter to balance the trade-off between the residual noise power and its distribution. To demonstrate the importance of $\lambda$, we report the simulation results in Fig. \ref{fig:necessity_norm_test}, using different values of $\lambda$ in the enhanced BP-CNN decoder. Note that the simulation uses the basic setting except the value of $\lambda$. To speed up the simulations, we use a low-complexity architecture of BP(5)-CNN-BP(5) in this set of experiments. We can see from Fig. \ref{fig:necessity_norm_test} that very small values of $\lambda$ cannot guarantee a Gaussian distribution for the residual noise, while very large $\lambda$ cannot depress the residual noise power. Hence a proper choice of hyperparameter $\lambda$ will affect the performance of BP-CNN. However, just like in many other deep learning tasks, optimizing $\lambda$ analytically is very difficult and we  select this parameter based on simulations. 

\begin{figure}
	\centering
	\includegraphics[width=0.8\linewidth]{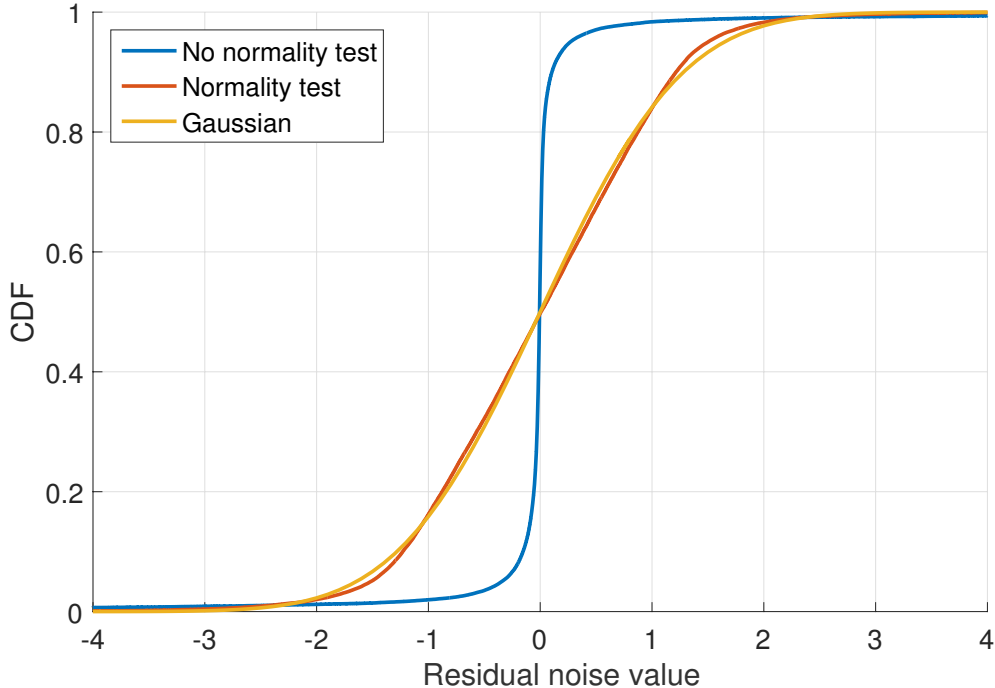}
	\caption{Compare the residual noise distribution with and without normality test. The residual noise is collected under $\eta=0.8$, SNR = $0$dB and the system structure is BP(5)-CNN-BP(5). When training the network with the normality test, $\lambda$ is set to be $0.1$.}
	\label{fig:res_noise_dist}
\end{figure}

Now let us take a deeper look at \eqref{eqn:loss_func} and numerically validate whether the normality test can indeed shape the output distribution to approximate Gaussian. We report the empirical cumulative distribution function (CDF) of the residual noise after CNN in Fig. \ref{fig:res_noise_dist}, both with (i.e., \emph{enhanced} BP-CNN) and without (i.e., \emph{baseline} BP-CNN) normality test. The residual noise data is collected under $\eta=0.8$, SNR = $0$dB and BP(5)-CNN-BP(5). When training the network with the normality test, $\lambda$ is set as 0.1. It is evident from Fig. \ref{fig:res_noise_dist} that involving the normality test in the loss function makes the residual noise distribution more like Gaussian.

From Fig. \ref{fig:necessity_norm_test}, we also find that for $\eta=0.8$, $\lambda=0.1$ performs the best among all tested $\lambda$ values, while $\lambda=10$ performs the best for $\eta=0.5$. We observe that a smaller $\lambda$ is preferred when $\eta=0.8$ compared to $\eta=0.5$, which can be explained as follows. With strong correlation in the channel noise, more information can be utilized to remove the errors of the network input and thus a smaller $\lambda$ is preferred to make the network focus on reducing the residual noise power. On the other hand, when correlation becomes weak and input elements are more independent, less information can be utilized to remove errors. In this case, the network needs to pay more attention to the distribution of the residual noise and a larger $\lambda$ is thus favored.

\subsubsection{Multiple iterations between CNN and BP further improve the performance}

\begin{figure}
	\centering
	\subfigure[$\eta=0.8$, strong correlation.]{ \includegraphics[width=0.6\linewidth]{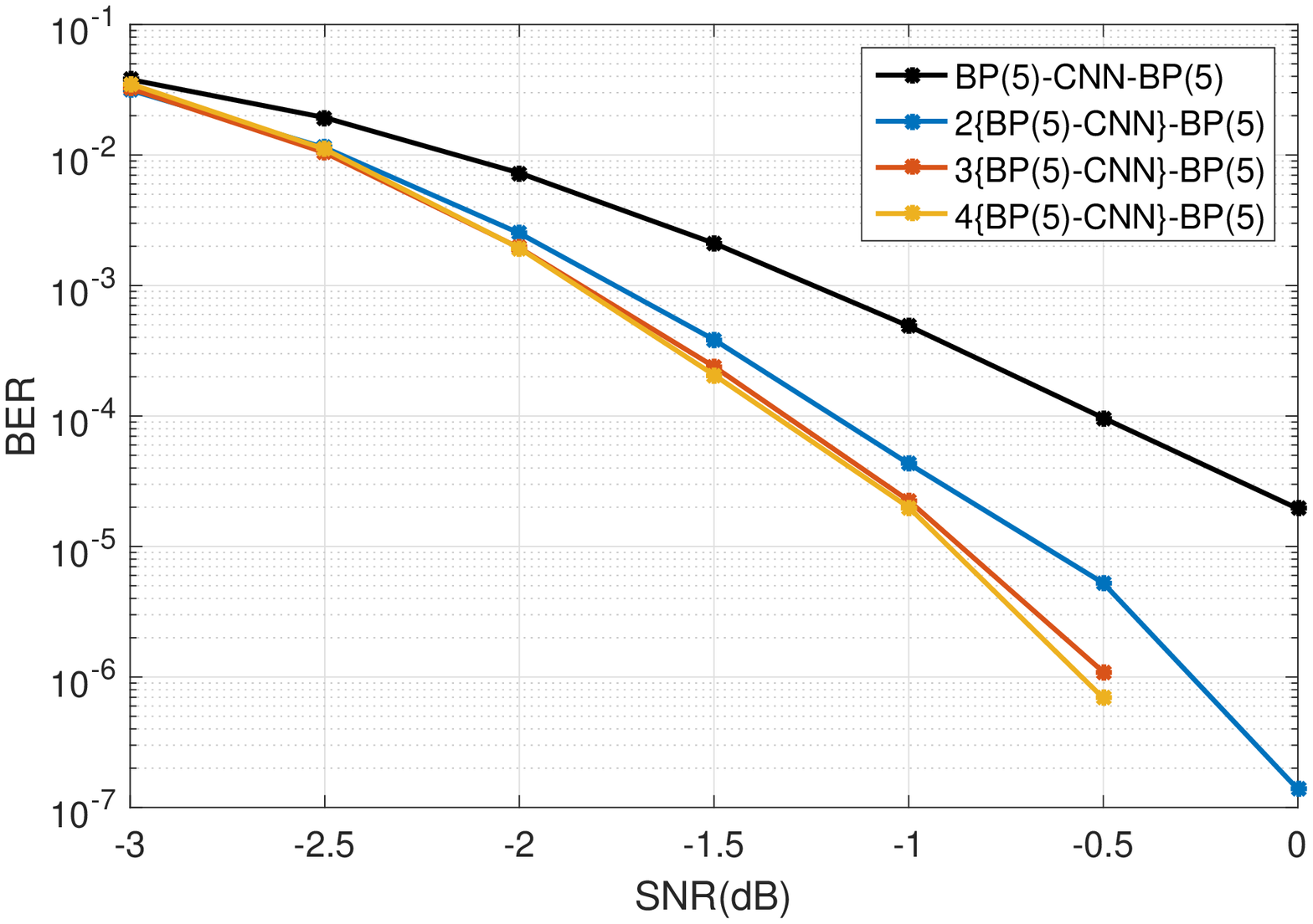}}
	\subfigure[$\eta=0.5$, moderate correlation.]{ \includegraphics[width=0.6\linewidth]{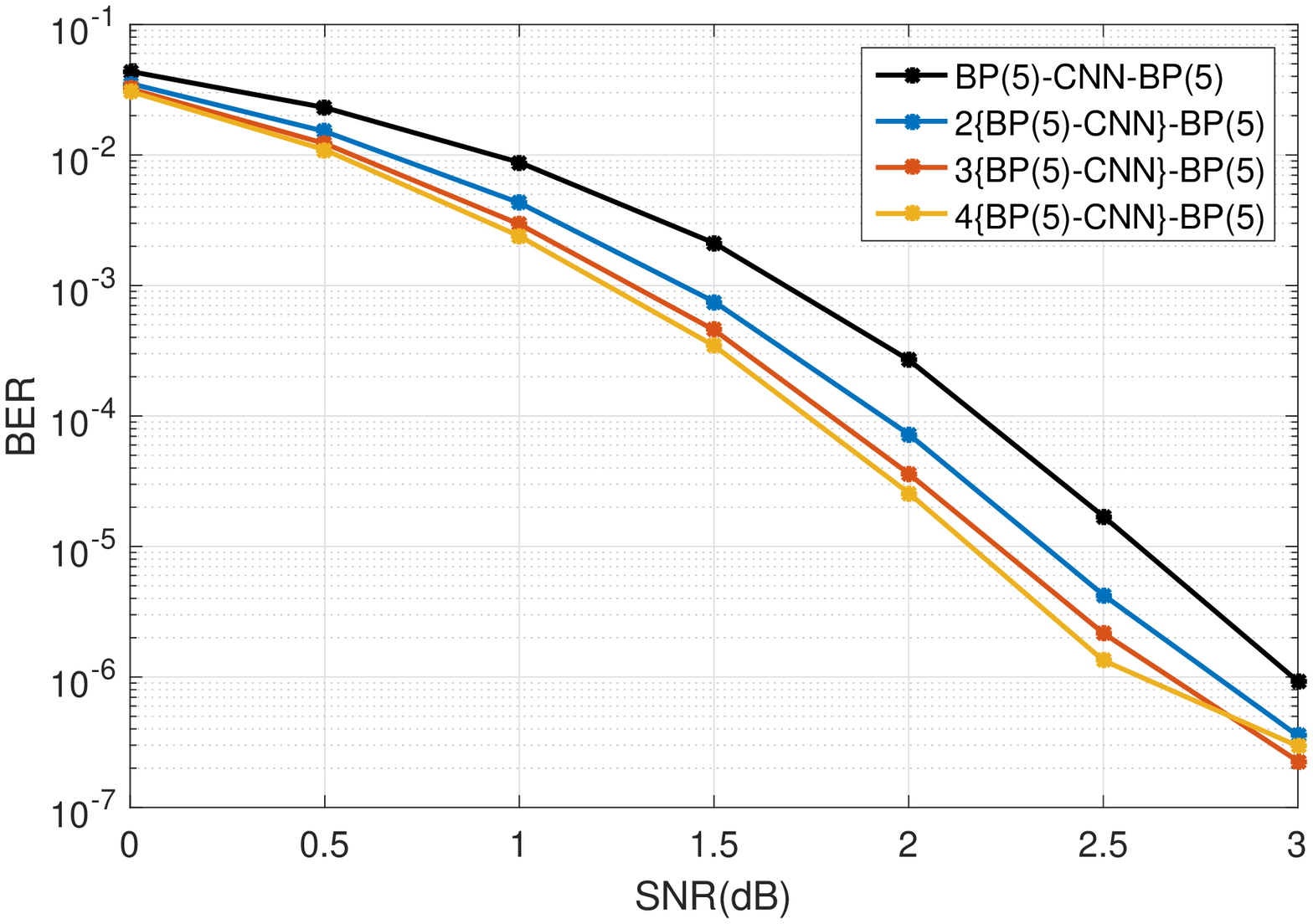}}
	\caption{Multiple iterations between CNN and BP can further improve decoding performance.}
	\label{fig:iterative_bp_cnn}
\end{figure}
Up to this point, we have only presented simulation results of the proposed iterative BP-CNN decoder with one iteration. Naturally, we can perform multiple iterations between CNN and BP with the hope of further reducing the BER, as shown in Fig. \ref{fig:sys_framework}. Notation-wise, we use $K$\{BP($n$)-CNN\}-BP($n$) to denote the iterative BP-CNN decoder structure with $K$ iterations between BP and CNN, and $n$ iterations inside BP. In total, $K+1$ BP($n$) and $K$ CNN are run. {Enhanced} BP-CNN is adopted for this set of experiments. We report the simulation results with different $K$'s in Fig. \ref{fig:iterative_bp_cnn}. It is clear that multiple iterations  further improve the decoding performance. When $\eta=0.8$, compared to the receiver architecture with only one iteration, two BP-CNN iterations improves the decoding performance by 0.7dB at BER=$10^{-4}$. In addition, we notice that after four BP-CNN iterations, the performance improvement becomes  insignificant. This is because the CNN has reached its maximum capacity and cannot further depress the residual noise power.

\subsubsection{BP-CNN is robust under different correlation models}
\begin{figure}
	\centering
	\includegraphics[width=0.8\linewidth]{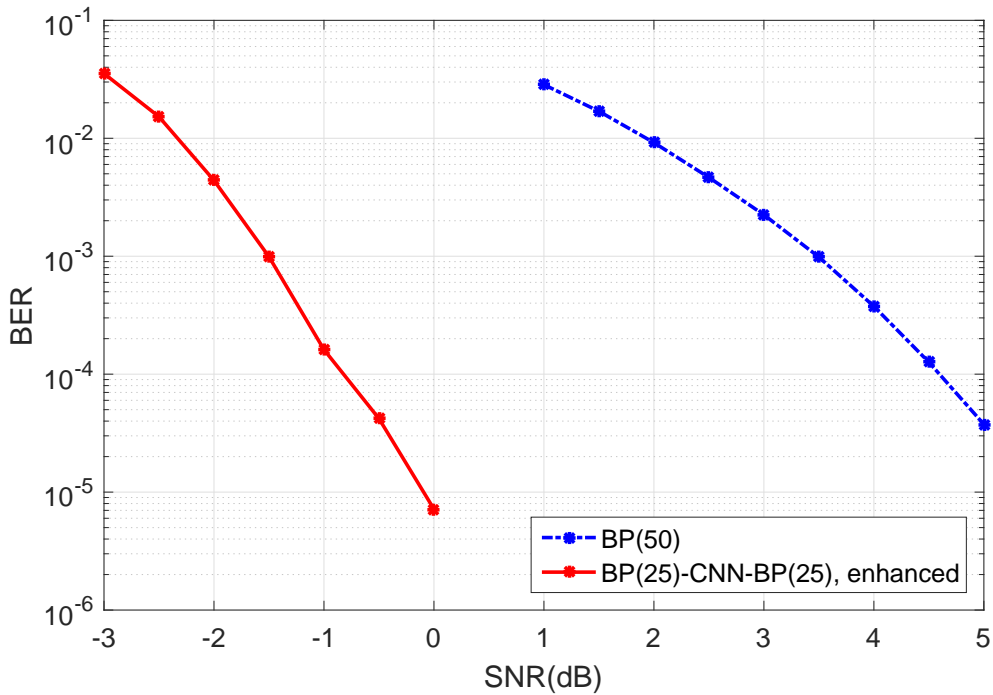}
	\caption{Performance test with pink noise. $\lambda=0.1$.}
	\label{fig:test_pink_noise}
\end{figure}

So far all the simulation results are  obtained using the noise correlation model defined in \eqref{eqn:corr_model}. As mentioned in Section~\ref{sec:system_framework}, the proposed iterative BP-CNN decoder does not rely on this specific correlation model. This should be intuitively reasonable as neither BP nor CNN uses the specific format of \eqref{eqn:corr_model}.  We now verify this statement by simulating the iterative BP-CNN decoder with another correlation model whose characteristics are described by the following power spectrum density:
\begin{equation}
\label{eqn:corr_model_psd}
P(f)\propto 1/|f|^\alpha.
\end{equation}
When $\alpha=1$, the noise is commonly referred to as \textit{pink noise}. 

We simulate the enhanced BP-CNN with pink noise and the results are reported in Fig. \ref{fig:test_pink_noise}. $\lambda$ is set to 0.1 for training the network. Clearly, the proposed method still achieves significant performance gains, proving that the proposed method can support different correlation models.

\subsection{BP vs. CNN: Impact on the Decoding Performance}

In the proposed framework, both the BP decoder and the CNN contribute to the improved decoding performance, but in very different ways. The BP decoder estimates the transmitted bits based on the encoding structure, while the CNN depresses the noise  by utilizing its  correlation. An important question can be asked: for a given amount of computation resources, how should the designer allocate the system complexity between BP and CNN, to obtain the best decoding performance?

A complete answer to this question remains unknown, and we resort to numerical simulations to provide some insight. In the simulations, we alter the complexity assignment between BP and CNN, and track the performance changes. We focus on the simplest form with only one iteration between  BP  and  CNN, so that the receiver structure is concisely denoted as BP-CNN-BP. We start our investigation from a structure with relatively low complexity, and add complexity to the BP decoder and the CNN respectively to observe how the performance improves. 

\begin{figure}
	\centering
	\subfigure[$\eta=0.8$, strong correlation.]{ \includegraphics[width=0.6\linewidth]{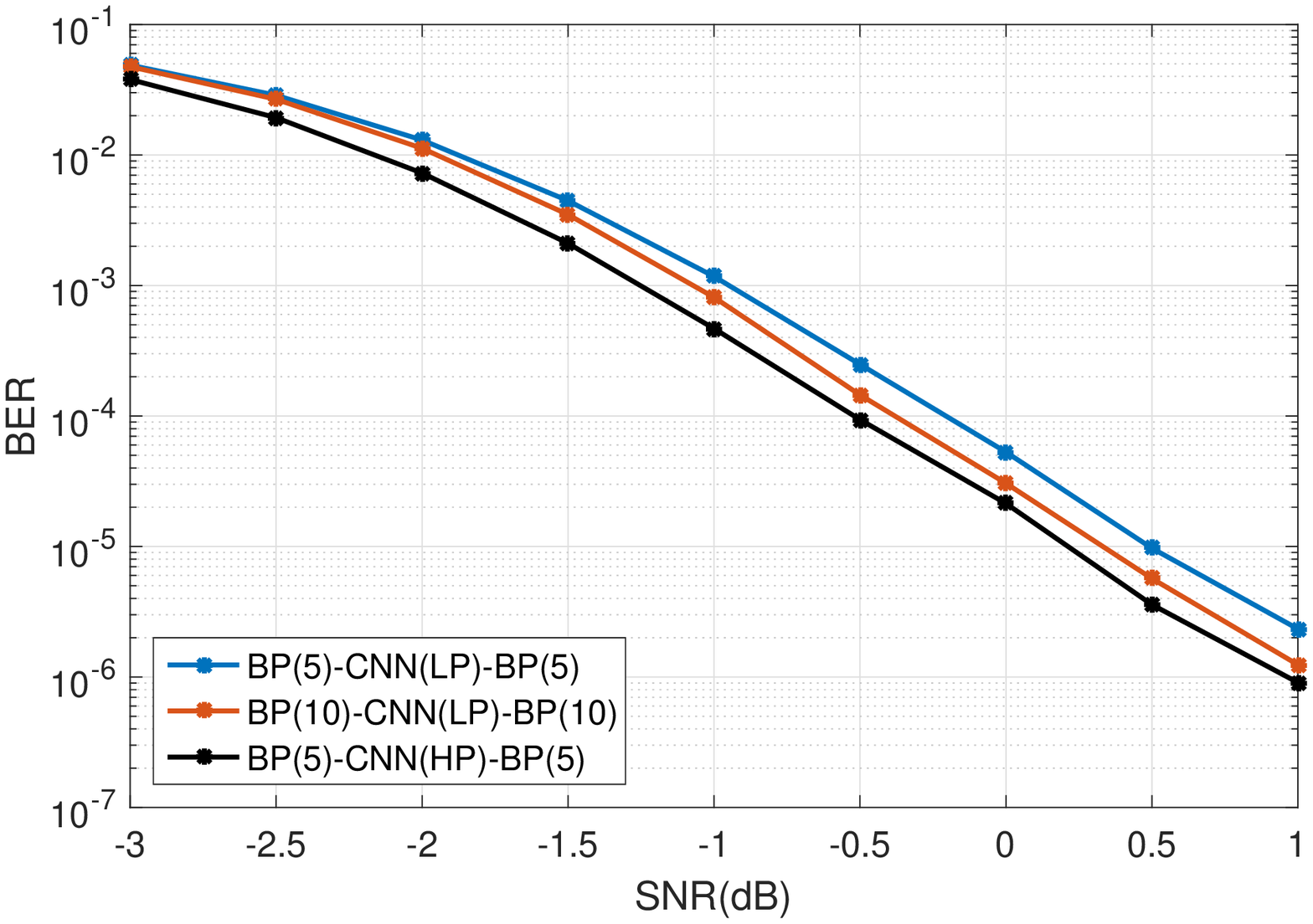}}
	\subfigure[$\eta=0.5$, moderate correlation.]{ \includegraphics[width=0.6\linewidth]{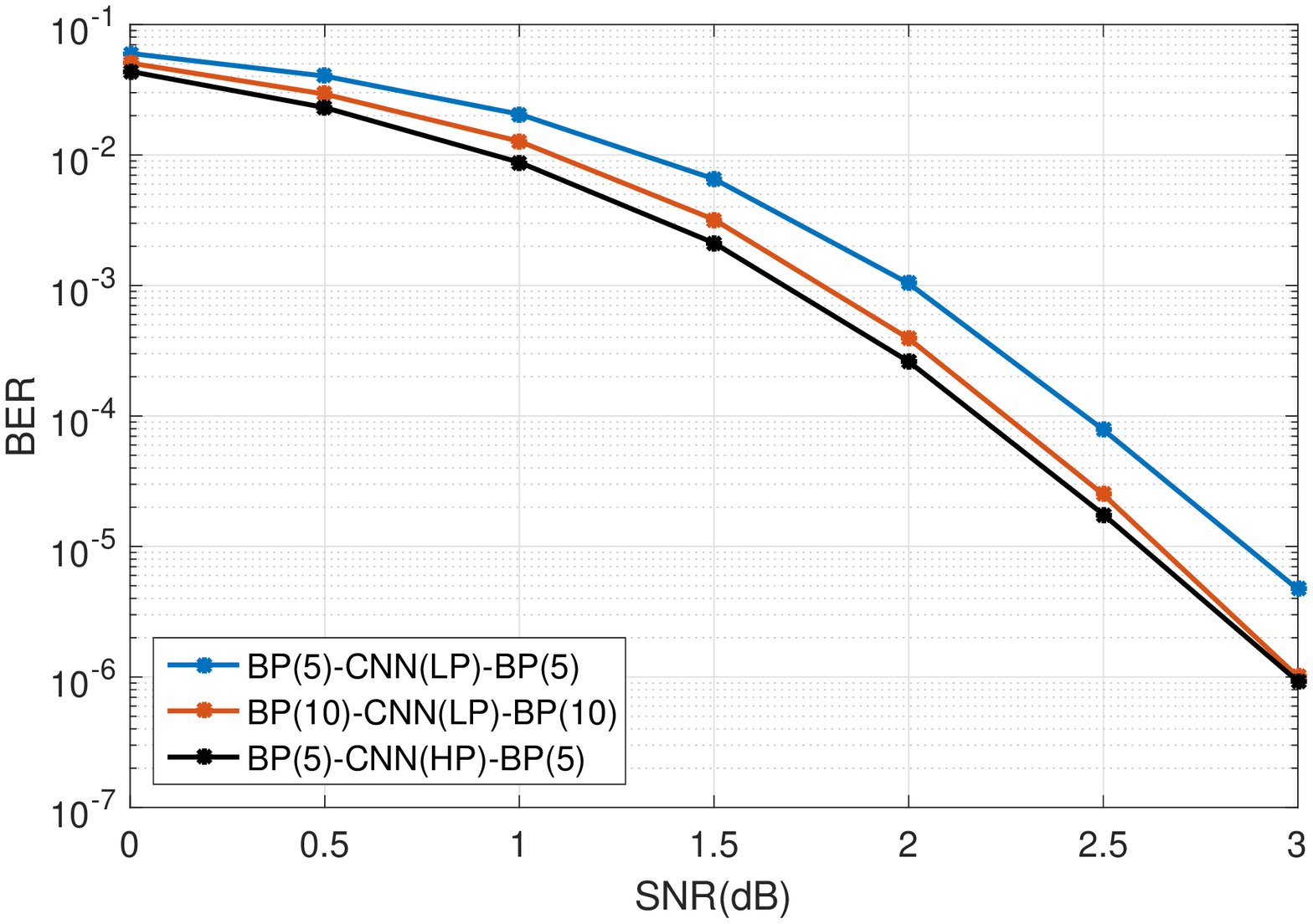}}
	\caption{The influences of different modules to the system performance. The legend CNN(LP) denotes a low-complexity network structure $\{3;5,1,9;16,8,1\}$ and CNN(HP) denotes a high-complexity network structure $\{4;9,3,3,15;64,32,16,1\}$.}
	\label{fig:analyze_modules}
\end{figure}

The results for $\eta=0.8$ and $\eta=0.5$ are given in Fig. \ref{fig:analyze_modules}. In each case, we first test a relatively low-complexity structure, denoted as BP(5)-CNN(LP)-BP(5), with five BP iterations executed in each BP decoding process. ``CNN(LP)'' denotes a low-complexity CNN, with its structure defined as $\{3;5,1,9;16,8,1\}$. Next, we add another five BP iterations to each BP decoding process to test a relatively high-complexity (with respect to BP) structure BP(10)-CNN(LP)-BP(10). Finally, we adopt a CNN structure $\{4;9,3,3,15;64,32,16,1\}$ (denoted as ``CNN(HP)''), and test a relatively high-complexity (with respect to CNN) structure BP(5)-CNN(HP)-BP(5). First, it is clear that  increasing the complexity of the BP decoder can improve the system performance, when the underline CNN structure is unchanged. Second, as  mentioned before, the complexity of the CNN(HP) structure is roughly equivalent to two BP iterations. Thus the receiver structure of BP(5)-CNN(HP)-BP(5) has lower complexity than BP(10)-CNN(LP)-BP(10). However, as observed from Fig. \ref{fig:analyze_modules}, BP(5)-CNN(HP)-BP(5) yields a better performance than BP(10)-CNN(LP)-BP(10). This demonstrates the necessity of introducing a CNN for decoding when the channel noise has strong or moderate correlation. The BP algorithm does not have the capability to extract the features existed in the channel noise for decoding. Therefore, when strong correlation exists, it is more effective to increase the CNN complexity to learn the channel characteristics  than to increase the complexity of the BP decoder.

\begin{figure}
	\centering
	\includegraphics[width=0.8\linewidth]{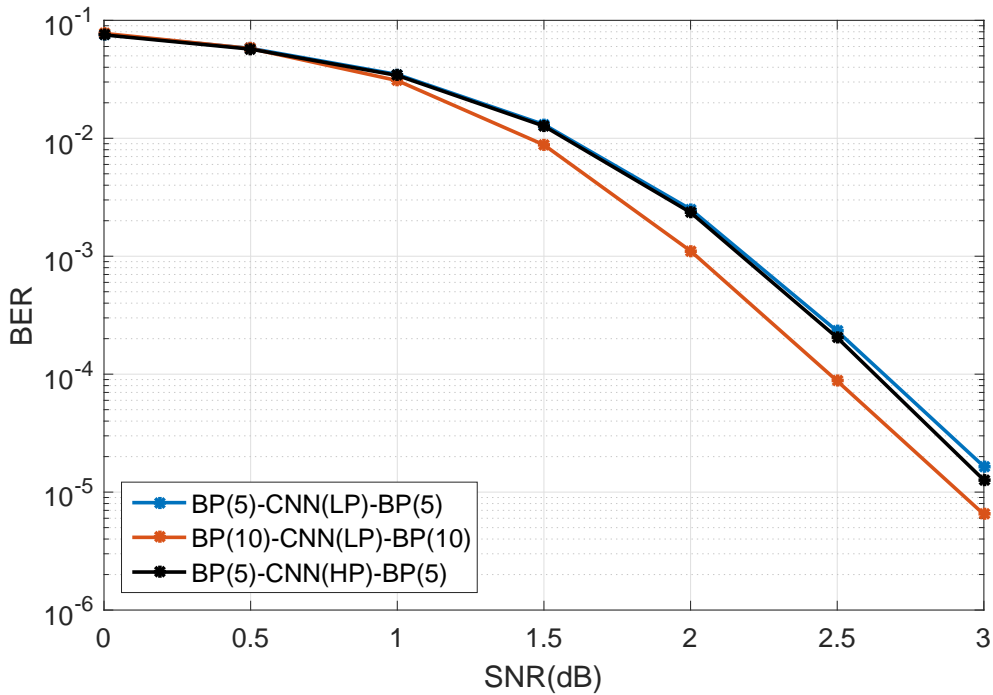}
	\caption{Experiments under AWGN channels (i.e., $\eta=0$) to show how the BP decoder and the CNN affect the system performance. The legend CNN(LP) denotes a low-complexity network structure $\{3;5,1,9;16,8,1\}$ and CNN(HP) denotes a high-complexity network structure $\{4;9,3,3,15;64,32,16,1\}$.}
	\label{fig:analyze_modules_awgn}
\end{figure}

However, the above statement is not valid when the correlation is weak enough so that  very few characteristics of the channel noise can be extracted by the CNN. To verify this, we report a set of tests under the  AWGN channel in Fig. \ref{fig:analyze_modules_awgn}. Intuitively, in the AWGN case where channel has no ``feature'' for the CNN to extract, assigning more complexity to the BP decoder is more effective.  This is verified in Fig. \ref{fig:analyze_modules_awgn}.

\subsection{The Impact of Training Data Generated under Different Channel Conditions}

\begin{figure}
	\centering
	\subfigure[$\eta=0.8$, strong correlation. $\lambda=0.1$.]{ \includegraphics[width=0.6\linewidth]{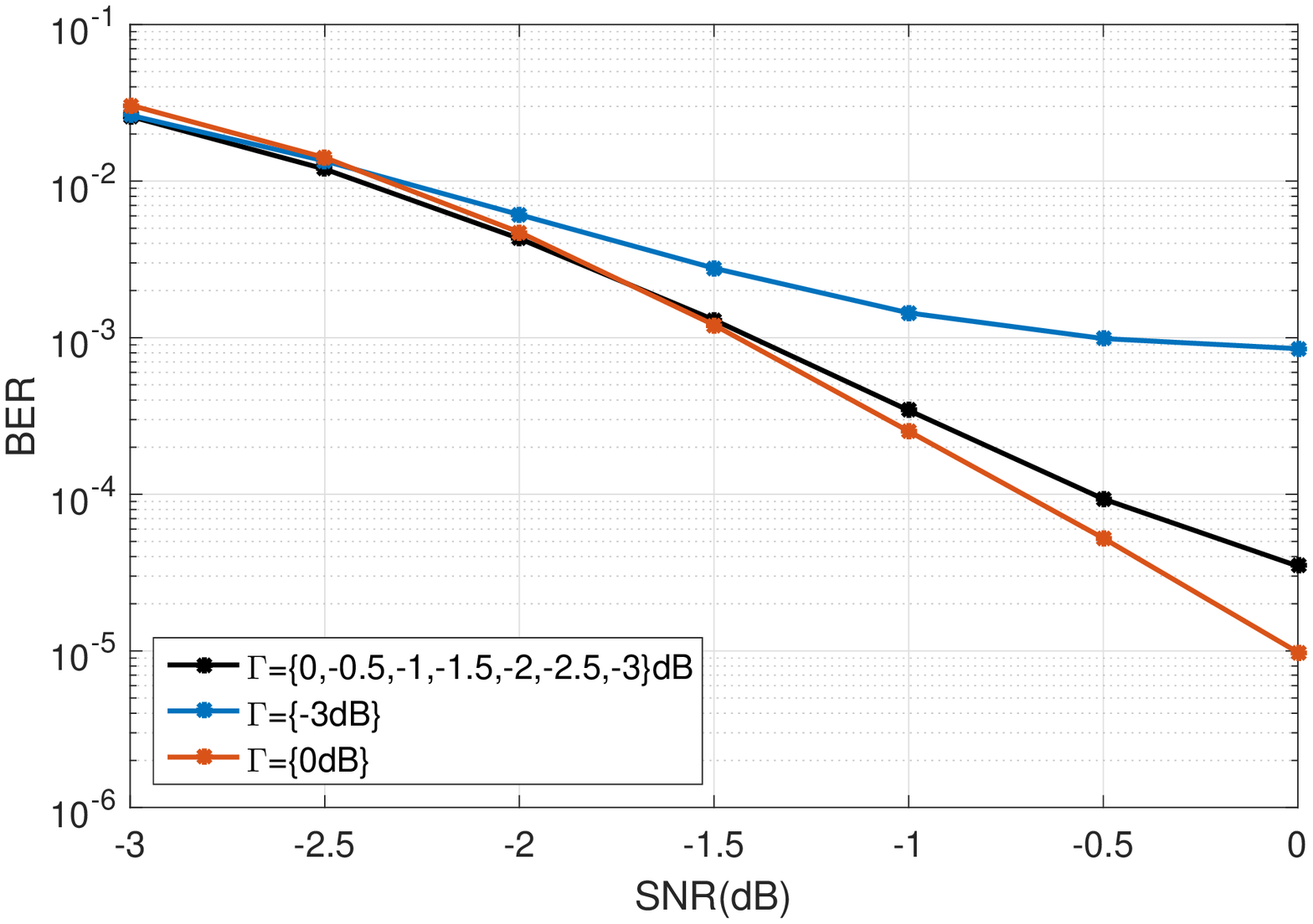} \label{fig:robust_snr1}}
	\subfigure[$\eta=0.5$, moderate correlation. $\lambda=10$.]{ \includegraphics[width=0.6\linewidth]{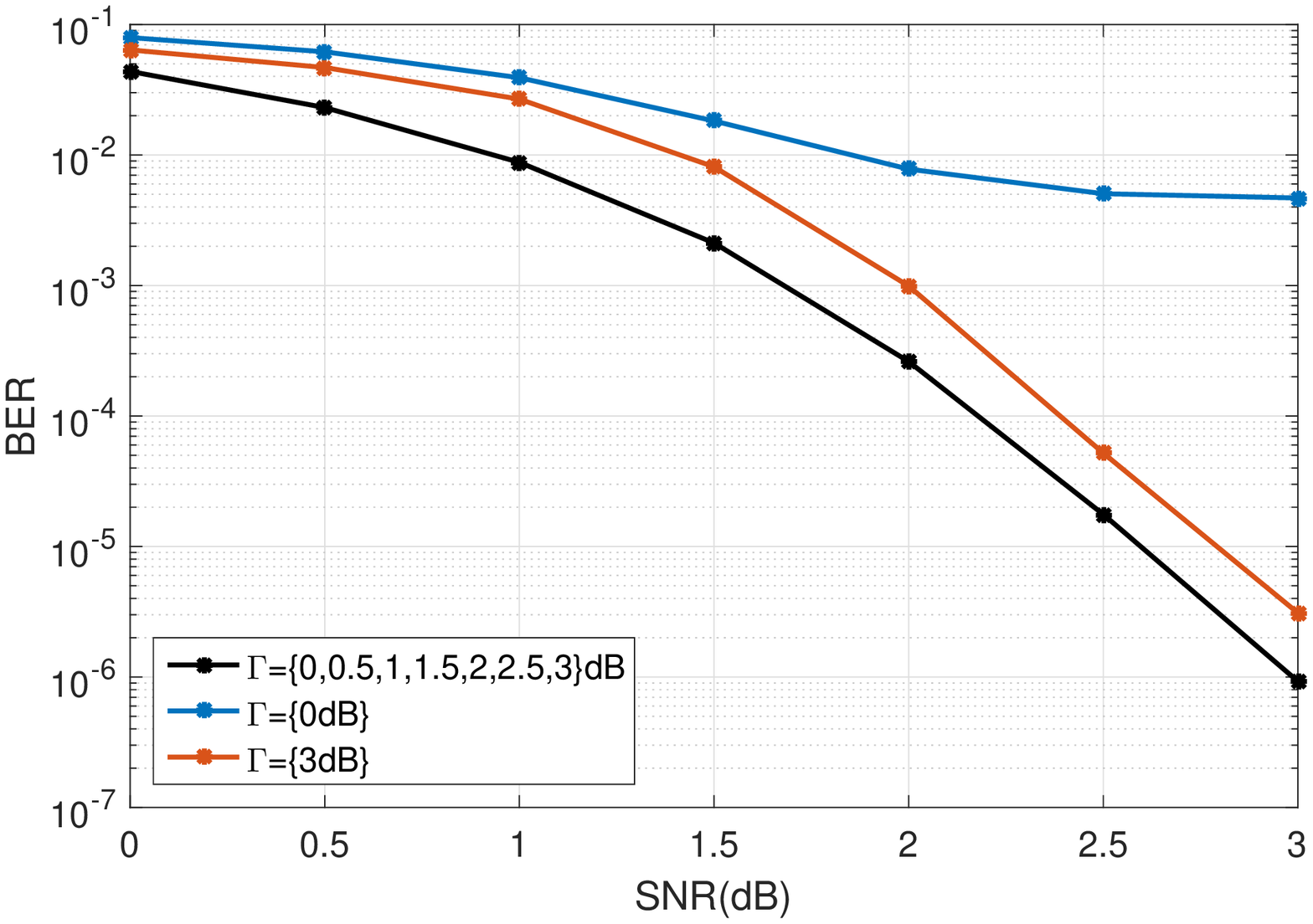} \label{fig:robust_snr2}}
	\caption{Analyze the system robustness to the training data generated under different channel conditions.}
	\label{fig:robust_snr}
\end{figure}

In the basic setting of our experiments, the training data is generated under multiple channel SNRs, i.e. $\Gamma=\{0,0.5,1,1.5,2,2.5,3\}$dB. In a mini-batch, each channel SNR occupies the same proportion of the training data. In practice, the operational range of SNR may not be known a priori, and hence training SNR and operation SNR may have a mismatch. Thus, in this section, we focus on training the network with data generated under a \textit{single} SNR and investigate how robust the BP-CNN decoder is under a range of SNR's. 

It is worth noting that in the previous results, we  have already reported some results with mismatched SNRs. For example, when $\eta=0.8$, the channel SNR range for evaluating the performance, i.e. $-3\sim1$dB, is already different from the SNR range for training data. This is because the strong correlation exists in the noise and CNN can achieve significant performance gains, and we have tested the BER performance under channel SNRs that are lower than the training SNRs. In this section, we report a more complete set of results.

In order to analyze how the data generation influences the system performance, we change the channel SNR range in the following experiments. Specifically, when $\eta=0.8$ we conduct the experiments under three different settings: $\Gamma=\{-3,-2.5,-2,-1.5,-1,-0.5,0\}$dB, $\Gamma=\{-3\}$dB and $\Gamma=\{0\}$dB. When $\eta=0.5$, we consider $\Gamma=\{0,0.5,1,1.5,2,2.5,3\}$dB, $\Gamma=\{0\}$dB and $\Gamma=\{3\}$dB. The decoder architecture is BP(5)-CNN-BP(5) and the CNN structure is $\{4;9,3,3,15;64,32,16,1\}$.

The results are reported in Fig. \ref{fig:robust_snr}. Generally speaking, generating training data under a wide range of channel SNRs is a better choice. As shown in Fig.~\ref{fig:robust_snr2}, training the network for 0dB or 3dB yields a worse performance than $\{0,0.5,1,1.5,2,2.5,3\}$dB. This is mainly ascribed to deep learning technologies which depend on training data to select network parameters. When $\Gamma=\{3\}$dB, the SNR is good and there are few errors in the input to CNN, i.e. $\hat{\mathbf{n}}$. With these ``skewed'' data, it is difficult for the network to learn how to remove errors. On the other hand, it is also not a good choice to generate training data under a bad channel condition, such as $\Gamma=\{0\}$dB in Fig.~\ref{fig:robust_snr2}. In this case, the training data contains too many errors, which is also not beneficial for the network to learn robust features of the channel noise.

Similar results are also obtained for $\eta=0.8$ as shown in Fig.~\ref{fig:robust_snr2}. It is worth noting that generating data under 0dB provides a similar performance as generating data under -3 to 0dB. It indicates that generating data under 0dB is also a good choice when $\eta=0.8$. Ultimately, how to select the \textit{optimal} channel conditions for data generation is a difficult task. Therefore, we suggest that the training data be generated under multiple channel conditions to enhance the data diversity. 

\section{Conclusions and future directions}
\label{sec:conclusion}

In this paper, we have designed a novel iterative BP-CNN  decoding structure to handle correlated channel noise.  The proposed framework serially concatenates a CNN with a BP decoder, and iterates between them. The BP decoder is to estimate the coded bits and indirectly estimate the channel noise. The CNN is to remove the channel noise estimation errors of the BP decoder by learning the noise correlation. To implement the framework, we have proposed to adopt a fully convolutional network structure and provided two strategies to train the network. We have carried out extensive simulations to show the effectiveness of the proposed iterative BP-CNN  decoder.

There are some important research directions for the iterative BP-CNN framework, which is worth investigating in the future. First, the loss function is not completely equivalent to the system performance measurement, and finding a better loss function is one of the important future directions. Furthermore, as mentioned in Section \ref{sec:system_summary}, the iterative system proposed in this paper can be unfolded into an open-loop system, and for maximum flexibility, we can design different number of BP iterations in each BP decoding process as well as different CNN structures. This system can be denoted as BP$(n_1)$-CNN1-BP$(n_2)$-CNN2-...-BP$(n_x)$-CNN$x$-BP$(n_{x+1})$, and provides the maximum degrees of freedom as we can allocate the complexity in all components to maximize the performance. We plan to study this general architecture in a future work.

\bibliographystyle{IEEEtran}
\bibliography{bc}

\end{document}